\documentclass[11pt,a4paper]{article}
\usepackage[hyperref]{emnlp2020}
\usepackage{times}
\usepackage{latexsym}
\usepackage{color}
\usepackage{graphicx}
\usepackage{amsmath}
\usepackage{amssymb}
\usepackage{caption}
\usepackage{algorithm}
\usepackage{algorithmicx}
\usepackage{algpseudocode}
\usepackage{multicol}
\usepackage{multirow}
\usepackage{booktabs}
\usepackage{adjustbox}
\usepackage{longtable}
\usepackage{listings}
\usepackage{url}
\usepackage{bm}
\usepackage{xspace}
\usepackage{makecell}
\usepackage{lipsum}
\usepackage{comment}
\usepackage{subcaption}

\usepackage{microtype}

\aclfinalcopy 
\def\aclpaperid{2048} 

\title{MAVEN: A Massive General Domain Event Detection Dataset}

\author{Xiaozhi Wang$^{1}$, Ziqi Wang$^{1}$, Xu Han$^{1}$, Wangyi Jiang$^{1}$, Rong Han$^{1}$, \\ \textbf{Zhiyuan Liu}$^{1,2}$\thanks{\quad Corresponding author: Z.Liu (liuzy@tsinghua.edu.cn)}\hspace{0.5em}, \textbf{Juanzi Li}$^{1,2}$, \textbf{Peng Li}$^{3}$, \textbf{Yankai Lin}$^{3}$, \textbf{Jie Zhou}$^{3}$\\
$^{1}$Department of Computer Science and Technology, BNRist;\\
$^{2}$KIRC, Institute for Artificial Intelligence, \\ Tsinghua University, Beijing, 100084, China\\
$^{3}$Pattern Recognition Center, WeChat AI, Tencent Inc, China\\
\texttt{\{wangxz20, ziqi-wan16, hanxu17\}@mails.tsinghua.edu.cn}\\ 
 }

\date{}

\begin{document}
\maketitle

\newcommand\blfootnote[1]{%
\begingroup
\renewcommand\thefootnote{}\footnote{#1}%
\addtocounter{footnote}{-1}%
\endgroup
}

\begin{abstract}

Event detection (ED), which means identifying event trigger words and classifying event types, is the first and most fundamental step for extracting event knowledge from plain text. Most existing datasets exhibit the following issues that limit further development of ED: (1) \textbf{Data scarcity}. Existing small-scale datasets are not sufficient for training and stably benchmarking increasingly sophisticated modern neural methods. (2) \textbf{Low coverage}. Limited event types of existing datasets cannot well cover general-domain events, which restricts the applications of ED models. To alleviate these problems, we present a MAssive eVENt detection dataset (MAVEN), which contains $4,480$ Wikipedia documents, $118,732$ event mention instances, and $168$ event types. MAVEN alleviates the data scarcity problem and covers much more general event types. We reproduce the recent state-of-the-art ED models and conduct a thorough evaluation on MAVEN. The experimental results show that existing ED methods cannot achieve promising results on MAVEN as on the small datasets, which suggests that ED in the real world remains a challenging task and requires further research efforts. We also discuss further directions for general domain ED with empirical analyses. The source code and dataset can be obtained from \url{https://github.com/THU-KEG/MAVEN-dataset}. 


\end{abstract}

\section{Introduction}

Event detection (ED) is an important task of information extraction, which aims to identify event triggers (the words or phrases evoking events in text) and classify event types. For instance, in the sentence ``\textit{Bill Gates \texttt{founded} Microsoft in 1975}'', an ED model should recognize that the word ``\textit{\texttt{founded}}'' is the trigger of a \texttt{Found} event. ED is the first stage to extract event knowledge from text ~\cite{ahn2006stages} and also fundamental to various NLP applications~\cite{yang2003structured,dart2014wsBasile,cheng2018implicit,yang2019interpretable}. 

\begin{figure}[t]
\centering
\includegraphics[width = 0.48\textwidth]{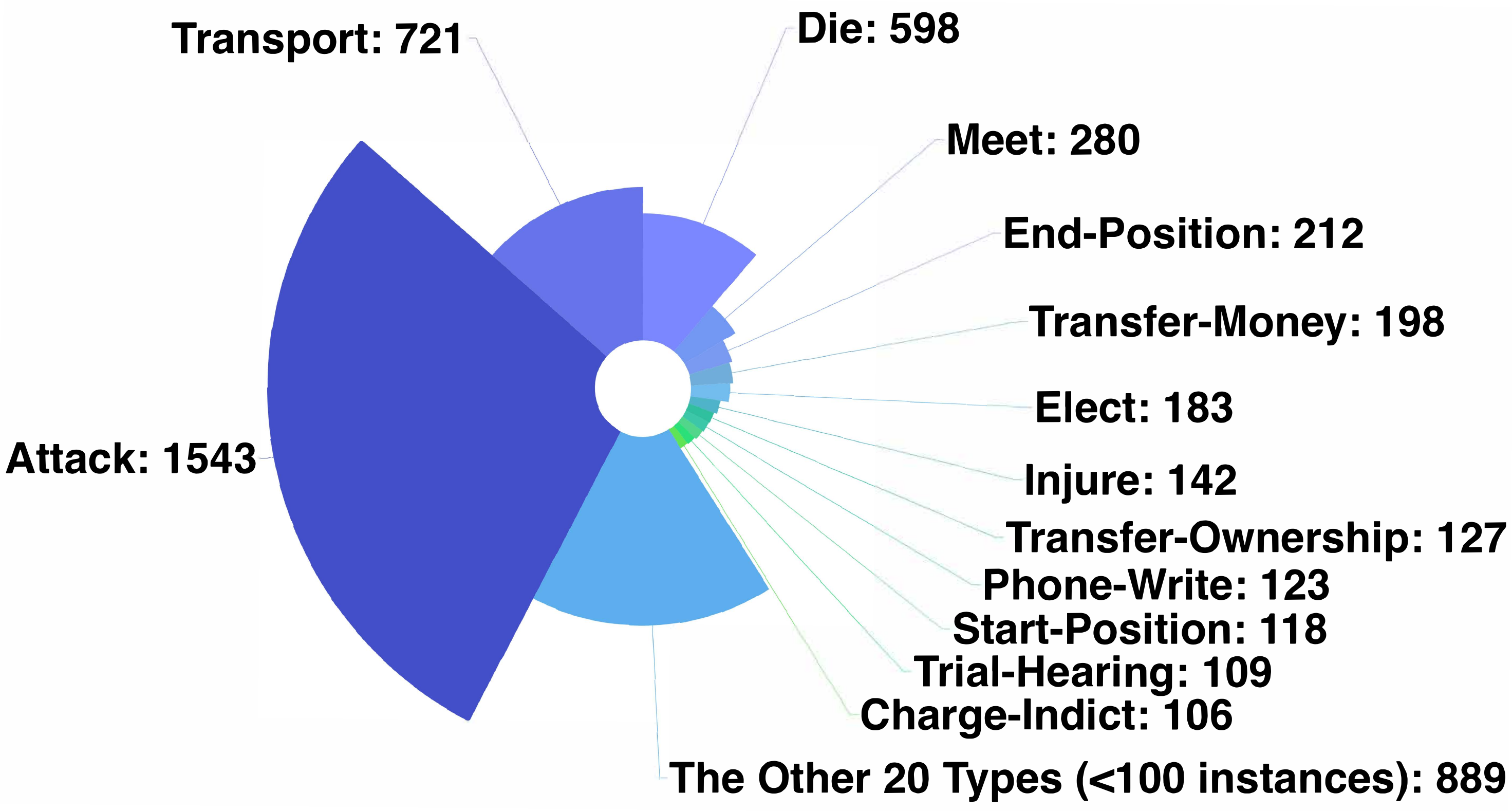}
\caption{Data distribution of the most widely-used ACE 2005 English dataset. It contains $33$ event types, $599$ documents and $5,349$ instances in total. }
\label{fig:ACEstat}
\end{figure}


Due to the rising requirement of event understanding, many efforts have been devoted to ED in recent years. The advanced models have been continuously proposed, including the feature-based models~\cite{ji2008refining,gupta-ji:2009:Short,li2013joint,araki2015joint} and advanced neural models~\cite{chen2015event,nguyen-grishman-2015-event,nguyen2016joint,feng2016language,ghaeini2016event,liu-etal-2017-exploiting,zhao-etal-2018-document,chen-etal-2018-collective,ding-etal-2019-event,yan-etal-2019-event}. Nevertheless, the benchmark datasets for ED are upgraded slowly. As event annotation is complex and expensive, the existing datasets are mostly small-scale. As shown in Figure~\ref{fig:ACEstat}, the most widely-used ACE 2005 English dataset~\cite{walker2006ace} only contains $599$ documents and $5,349$ annotated instances. Due to the inherent data imbalance problem, $20$ of its $33$ event types only have fewer than $100$ annotated instances. As recent neural methods are typically data-hungry, these small-scale datasets are not sufficient for training and stably benchmarking modern sophisticated models. 
Moreover, the covered event types in existing datasets are limited. The ACE 2005 English dataset only contains $8$ event types and $33$ specific subtypes. The Rich ERE ontology~\cite{song-etal-2015-light} used by TAC KBP challenges~\cite{ellis2015overview,ellis2016overview} covers $9$ event types and $38$ subtypes. The coverage of these datasets is low for general domain events, which results in the models trained on these datasets cannot be easily transferred and applied on general applications. 

Recent research~\cite{huang-etal-2016-liberal,chen-etal-2017-automatically} has shown that the existing datasets suffering from the data scarcity and low coverage problems are now inadequate for benchmarking emerging methods, i.e., the evaluation results are difficult to reflect the effectiveness of novel methods. To tackle these issues, some works adopt the distantly supervised methods~\cite{mintz-etal-2009-distant} to automatically annotate data with existing event facts in knowledge bases~\cite{chen-etal-2017-automatically,zeng2018scale,araki-mitamura-2018-open} or use bootstrapping methods to generate new data~\cite{ferguson-etal-2018-semi,wang-etal-2019-adversarial-training}. However, the generated data are inevitably noisy and homogeneous due to the limited number and low diversity of event facts and seed data instances.

In this paper, we present MAVEN, a human-annotated massive general domain event detection dataset constructed from English Wikipedia 
and FrameNet~\cite{baker1998berkeley}, which can alleviate the data scarcity and low coverage problems: 

(1) Our MAVEN dataset contains $111,611$ different events, $118,732$ event mentions, which is twenty times larger than the most widely-used ACE 2005 dataset, and $4,480$ annotated documents in total. To the best of our knowledge, this is the largest human-annotated event detection dataset until now. 

(2) MAVEN contains $168$ event types, which covers a much broader range of general domain events. These event types are manually derived from the frames defined in the linguistic resource FrameNet~\cite{baker1998berkeley}, which has been shown to have good coverage of general event semantics~\cite{aguilar-etal-2014-comparison,huang-etal-2018-zero}. Furthermore, we construct a tree-structure hierarchical event type schema, which not only maintains the good coverage of FrameNet but also avoids the difficulty of crowd-sourced annotation caused by the original sophisticated schema, and may help future ED models with the hierarchy information.


We reproduce some recent state-of-the-art ED models and conduct a thorough evaluation of these models on MAVEN. From the experimental results, we observe significant performance drops of these models as compared with on existing ED benchmarks. It indicates that detecting general-domain events is still challenging and the existing datasets are difficult to support further explorations. We also explore some promising directions with empirical analyses, including modeling the multiple events shown in one sentence, using the hierarchical event schema to handle long-tail types and distinguish close types, and improving low-resource ED tasks 
with transfer learning. We hope that all contents of MAVEN could encourage the community to make further breakthroughs.

\section{Event Detection Definition}
\label{sec:taskDef}
In our dataset, we mostly follow the settings and terminologies defined in the ACE 2005 program~\cite{doddington2004automatic}. We specify the vital terminologies as follows:

An \textbf{event} is a specific occurrence involving participants~\cite{linguistic2005ace}. In MAVEN, we mainly focus on extracting the basic events that can be specified in one or a few sentences. Each event will be labeled with a certain \textbf{event type}. An \textbf{event mention} is a sentence within which the event is described. As the same event may be mentioned multiple times in a document, there are typically more event mentions than events. An \textbf{event trigger} is the key word or phrase in an event mention that most clearly expresses the event occurrence. 

The ED task is to identify event triggers and classify event types for given sentences. Accordingly, ED is conventionally divided into two subtasks: Trigger Identification and Trigger Classification~\cite{ahn2006stages}. \textbf{Trigger identification} is to identify the annotated triggers from all possible candidates. \textbf{Trigger classification} is to classify the corresponding event types for the identified triggers. Both the subtasks are evaluated with micro precision, recall, and F-1 scores. Recent neural methods typically formulate ED as a token-level multi-class classification task~\cite{chen2015event,nguyen2016joint} or a sequence labeling task~\cite{chen-etal-2018-collective,zeng2018scale}, and only report the trigger classification results (add an additional type \texttt{N/A} to be classified at the same time, indicating that the candidate is not a trigger). In MAVEN, we inherit all the above-mentioned settings in both dataset construction and model evaluation.


\section{Data Collection of MAVEN}
\label{sec:collection}


\subsection{Event Schema Construction}

The event schema used by the existing ED datasets like ACE~\cite{doddington2004automatic}, Light ERE~\cite{aguilar-etal-2014-comparison} and Rich ERE~\cite{song-etal-2015-light} only includes limited event types (e.g. \texttt{Movement}, \texttt{Contact}, etc). Hence, we need to construct a new event schema with a good coverage of general-domain events for our dataset.

Inspired by~\newcite{aguilar-etal-2014-comparison}, we mostly use the frames in FrameNet~\cite{baker1998berkeley} as our event types for a good coverage. FrameNet follows the frame semantic theory~\cite{fillmore1976frame,fillmore2006frame} and defines over $1,200$ semantic frames along with corresponding frame elements, frame relations, and lexical units. 
From the ED perspective, some frames and lexical units can be used as event types and triggers respectively.

Considering FrameNet is primarily a linguistic resource constructed by linguistic experts, it prioritizes lexicographic and linguistic completeness over ease of annotation~\cite{aguilar-etal-2014-comparison}. To facilitate the crowd-sourced annotation with large numbers of annotators, we simplify the original frame schema into our event schema.  
We collect $598$ event-related frames from FrameNet by recursively selecting the frames having ``Inheritance'', ``Subframe'' or ``Using'' relations with the \texttt{Event} frame like \newcite{Li2019JointEE}. Then we manually filter out abstractive frames (e.g. \texttt{Process\_resume}), merge similar frames (e.g. \texttt{Choosing} and \texttt{Adopt\_selection} ), and assemble too fine-grained frames into more generalized frames (e.g. \texttt{Visitor\_arrival} and \texttt{Drop\_in\_on} into \texttt{Arriving}). 
We finally get $168$ event types to annotate, covering $74.4\%$ (selected or inherit from the selected frames) of the $598$ event-related frames, and the mapping between event types and frames are shown in Appendix~\ref{app:frames}. 

Based on the FrameNet inheritance relation and the HowNet event schema~\cite{dong2003hownet}, we organize the event types into a tree-structure \textbf{hierarchical event type schema}. During annotation, we ask the annotators to label the triggers with the most fine-grained type (e.g. \texttt{Theft} and \texttt{Robbery}). The coarse-grained types (e.g. \texttt{Committing\_crime}) are only used for those rare events without appropriate fine-grained types so that to recall more events with fewer labels. Appendix~\ref{app:hierschema} shows the overall hierarchical schema.

\subsection{Document Selection}

To support the annotation, we need a large number of informative documents as our basic corpus. We adopt English Wikipedia as our data source considering it is informative and widely-used~\cite{rajpurkar-etal-2016-squad,yang-etal-2018-hotpotqa}. Meanwhile, Wikipedia articles contain rich entities, which will benefit event argument annotation in the future. 
\begin{table}
    \centering\small
        \begin{tabular}{lrr}
            \toprule
            \textbf{Topic}         & \textbf{\#Documents}    & \textbf{Percentage}    \\
            \midrule
            Military conflict   & $1,458$     & $32.5\%$        \\
            Hurricane           & $480$       & $10.7\%$        \\ 
            Civilian attack     & $287$       & $6.4\%$        \\
            Concert tour      & $255$       & $5.7\%$        \\
            Music festival    & $170$       & $3.8\%$        \\
            \midrule
            Total               & $2,650$    & $59.2\%$        \\
            \bottomrule
        \end{tabular}
    \caption{Count and \% of MAVEN documents in Top-5 EventWiki~\cite{ge-etal-2018-eventwiki} topics.}
    \label{tab:dataset_top_category}
\end{table}

To effectively select the articles containing enough events, we follow a simple intuition that the articles describing grand ``topic events'' may contain much more basic events than the articles about specific entity definitions. We adopt EventWiki~\cite{ge-etal-2018-eventwiki} to help select the event-related articles. It is a knowledge base for major events and each major event is described with a Wikipedia article. We thus utilize the articles indexed by EventWiki as the base and manually select some articles to annotate their basic events covered by our event schema. To ensure the quality of articles, we follow the previous settings~\cite{yao-etal-2019-docred} to use the introductory sections for annotation. Moreover, we filter out the articles with fewer than $5$ sentences or fewer than $10$ event-related frames labeled by a semantic labeling tool~\cite{swayamdipta2017frame}.

Finally, we select $4,480$ documents in total, covering $90$ of the $95$ major event topics defined in EventWiki. Table~\ref{tab:dataset_top_category} shows the top $5$ EventWiki topics of our selected documents.

\begin{table*}[t]
\centering
\scalebox{0.86}{
\begin{tabular}{clrrrrrr}
\toprule
\multicolumn{2}{c}{\textbf{Dataset}}                                                                                        & \textbf{\#Documents} & \textbf{\#Tokens} & \textbf{\#Sentences} & \textbf{\#Event Types} & \textbf{\#Events}  & \textbf{\#Event Mentions} \\ 
\midrule
\multicolumn{2}{c}{ACE 2005}                                                                                        & $599$       & $303$k & $15,789$  & $33$     & $4,090$   & $5,349$          \\ \midrule
\multicolumn{1}{c|}{\multirow{8}{*}{\begin{tabular}[c]{@{}c@{}}Rich\\ ERE\end{tabular}}} & LDC2015E29              &   $91$       & $43$k & $1,903$   & $38$          &  $1,439$   & $2,196$           \\
\multicolumn{1}{c|}{}                                                                    & LDC2015E68              &   $197$      & $164$k & $8,711$  & $37$          &  $2,650$   & $3,567$               \\
\multicolumn{1}{c|}{}                                                                    & LDC2015E78              &   $171$      & $114$k & $4,979$  & $31$          &  $2,285$   & $2,933$               \\
\multicolumn{1}{c|}{}                                                                    & TAC KBP 2014            &   $351$      & $282$k & $14,852$  & $34$          &  $10,719$  & $10,719$                \\
\multicolumn{1}{c|}{}                                                                    & TAC KBP 2015            &   $360$      & $238$k & $11,535$  & $38$          &  $7,460$    & $12,976$               \\
\multicolumn{1}{c|}{}                                                                    & TAC KBP 2016            &   $169$      & $109$k & $5,295$  & $18$          &  $3,191$    & $4,155$               \\
\multicolumn{1}{c|}{}                                                                    & TAC KBP 2017            &   $167$      & $99$k & $4,839$   & $18$          &  $2,963$    & $4,375$               \\ \cmidrule{2-8} 
\multicolumn{1}{c|}{}                                                                    & \multicolumn{1}{c}{Total} &   $1,272$     & $854$k & $41,708$  & $38$             &  $29,293$   & $38,853$               \\ \midrule
\multicolumn{2}{c}{MAVEN}                                                                                            & $\bm{4,480}$     & $\bm{1,276}$\textbf{k} & $\bm{49,873}$ & $\bm{168}$         & $\bm{111,611}$ & $\bm{118,732}$        \\ \bottomrule
\end{tabular}
}
     \caption{Statistics of MAVEN compared with existing widely-used ED datasets. The \#Event Type shows the number of the most fine-grained types (i.e. the ``subtype'' of ACE and ERE). For the multilingual datasets, we report the statistics of the English subset (typically the largest subset) for direct comparisons to MAVEN. We merge all the Rich ERE datasets and remove the duplicate documents to get the ``Total'' statistics. } 
     \label{tab:dataset_statistics}
\end{table*}

\subsection{Candidate Selection and Automatic Labeling}
\label{sec:candidate}

We have massive data to be annotated with $168$ event types. To facilitate efficiency and improve consistency of our annotators, who are not all linguistic experts, we adopt some heuristic methods to narrow down trigger candidates and the corresponding event type candidates, and automatically label some triggers to provide information.

\paragraph{Candidate selection }
We first do POS tagging with the NLTK toolkit~\cite{bird-2006-nltk}, and select the content words (nouns, verbs, adjectives, and adverbs) as the trigger candidates to be annotated. As event triggers can also be phrases, the phrases in documents that can be matched with the phrases provided in FrameNet are also selected as trigger candidates. For each trigger candidate, we provide $15$ event types as label candidates. The $15$ type candidates are automatically recommended with the cosine similarities between trigger word embeddings and the average of the word embeddings of event types' corresponding lexical units in FrameNet. The word embeddings we used here are the pre-trained Glove~\cite{pennington-etal-2014-glove} word vectors.
To verify the effectiveness of these candidate selection methods, we randomly choose $50$ documents and invite an expert to directly label all the words with the $168$ event types. The results show that $100\%$ of the expert-provided labeled triggers appeared among the automatically listed trigger candidates provided to annotators. Furthermore, the results also show that $96.8\%$ of the expert-provided event types appeared among the $15$ event type candidates automatically recommended to the annotators. 

\paragraph{Automatic labeling}
We label some trigger candidates with a state-of-the-art frame semantic parser~\cite{swayamdipta2017frame} and use the corresponding event types of the predicted frames as the default event types. The annotators can replace them with more appropriate event types or just keep them to save time and effort. Evaluated on the final dataset, the frame semantic parser can achieve $52.4\%$ precision and $49.7\%$ recall, which indicates that the automatic labeling process can help to save about a half of the overall annotation effort.

\subsection{Human Annotation}

The final step requires the annotators to label the trigger candidates with appropriate event types and merge the event mentions (annotate which mentions are expressing the same event).

\paragraph{Annotation process} As the event annotation is complicated, to ensure the accuracy and consistency of our annotation, we follow the ACE 2005 annotation process~\cite{linguistic2005ace} to organize a two-stage iterative annotation. In the first stage, $121$ crowd-source annotators are invited to annotate the documents given the default results and candidate sets described in the last section. Each document is annotated twice by two independent annotators in this stage. In the second stage, $17$ experienced annotators and experts will give the final results on top of the annotation results of the two first-stage annotators. Each document will be annotated only once in the second stage. 

\paragraph{Data quality} To evaluate the dataset quality, we randomly sample $1,000$ documents and invite different second-stage annotators to independently annotate these documents for one more time. We measure the inter-annotator agreements of the event type annotation between two annotators with Cohen's Kappa~\cite{cohen1960coefficient}. The results for the first stage trigger and type annotation are $38.2\%$ and $42.7\%$, respectively. And the results for the second stage trigger and type annotation are $64.1\%$ and $73.7\%$. One of the authors also manually examined $50$ random documents. The estimated accuracies of event type annotation and event mention merging are $90.1\%$ and $86.0\%$ respectively. These results show that although the general domain event annotation is difficult (the first-stage inter-agreement is low), MAVEN's quality is satisfactory.

\section{Data Analysis of MAVEN}


\begin{figure}[t]
\small
\centering
\scalebox{0.773}{
\includegraphics[width = 0.49\textwidth]{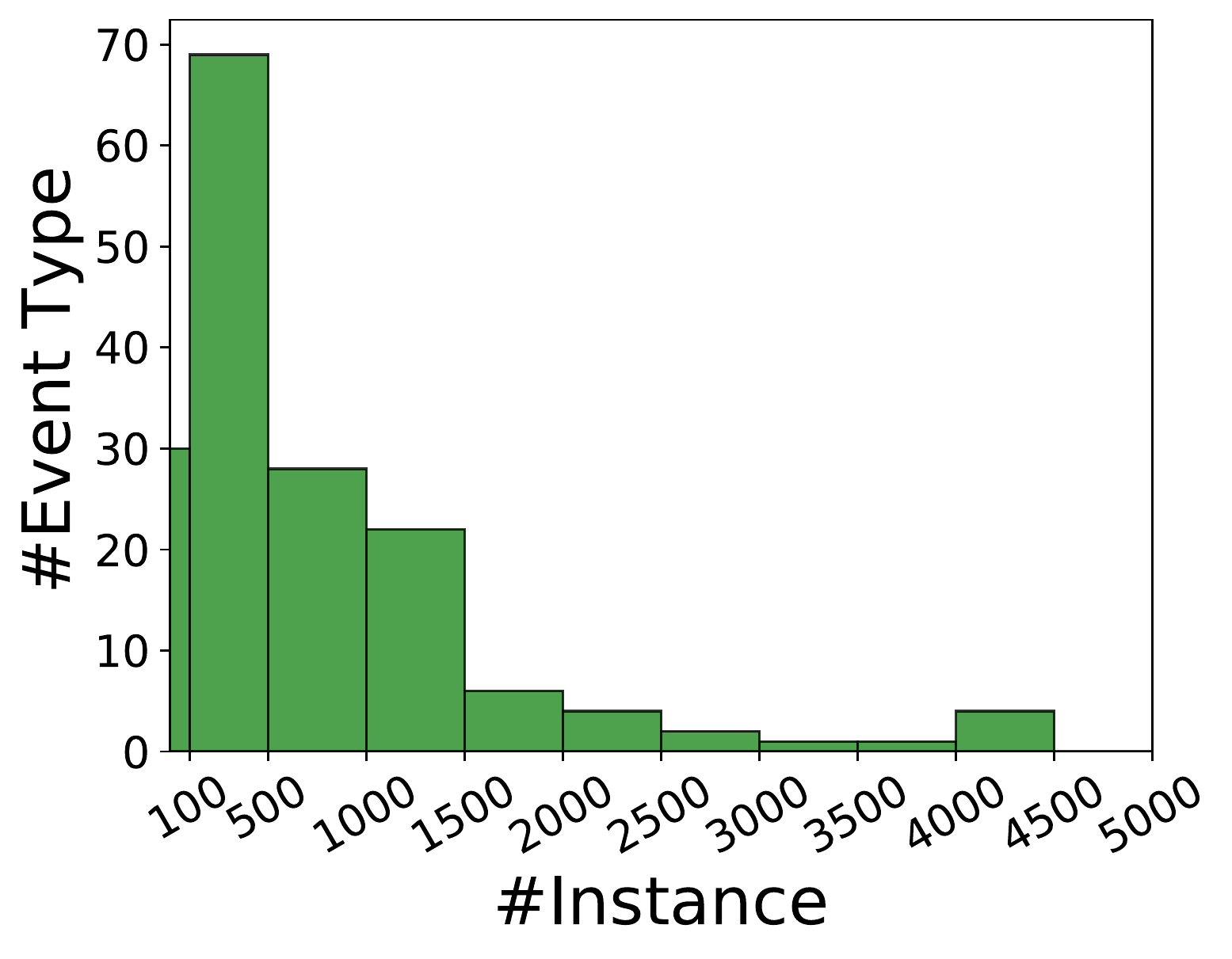}
}
\caption{Distribution of MAVEN event types by their instance numbers.}
\label{fig:ourDist}
\end{figure}

\begin{table}[tp]
\small
\centering
\scalebox{0.8}{
\begin{tabular}{l|c|c}
\toprule
\multicolumn{1}{c|}{\begin{tabular}[c]{@{}c@{}}\textbf{Top-level}\\ \textbf{Event Type}\end{tabular}} & \textbf{Subtype Examples}                                                                                         & \textbf{Percentage} \\ \midrule
\texttt{Action}                                                                              & \begin{tabular}[c]{@{}c@{}}\texttt{Telling}, \texttt{Attack},\\  \texttt{Building}\end{tabular}          & $46.9\%$   \\ \hline
\texttt{Change}                                                                              & \begin{tabular}[c]{@{}c@{}}\texttt{Change\_event\_time}, \\ \texttt{Change\_of\_leadership}\end{tabular} & $27.5\%$   \\ \hline
\texttt{Scenario}                                                                            & \begin{tabular}[c]{@{}c@{}}\texttt{Emergency}, \texttt{Catastrophe},\\  \texttt{Incident}\end{tabular}   & $13.4\%$   \\\hline
\texttt{Sentiment}                                                                           & \begin{tabular}[c]{@{}c@{}}\texttt{Supporting}, \texttt{Convincing}, \\ \texttt{Quarreling}\end{tabular} & $6.4\%$    \\\hline
\texttt{Possession}                                                                          & \begin{tabular}[c]{@{}c@{}}\texttt{Commerce\_buy}, \texttt{Giving}, \\ \texttt{Renting}\end{tabular}     & $5.7\%$    \\ \bottomrule
\end{tabular}
}
\caption{Five top-level event types and their percentages of MAVEN. Appendix~\ref{app:hierschema} shows more details.}
\label{tab:majortype}
\end{table}

\begin{figure}[t]
\small
\centering
\scalebox{0.8}{
\includegraphics[width = 0.49\textwidth]{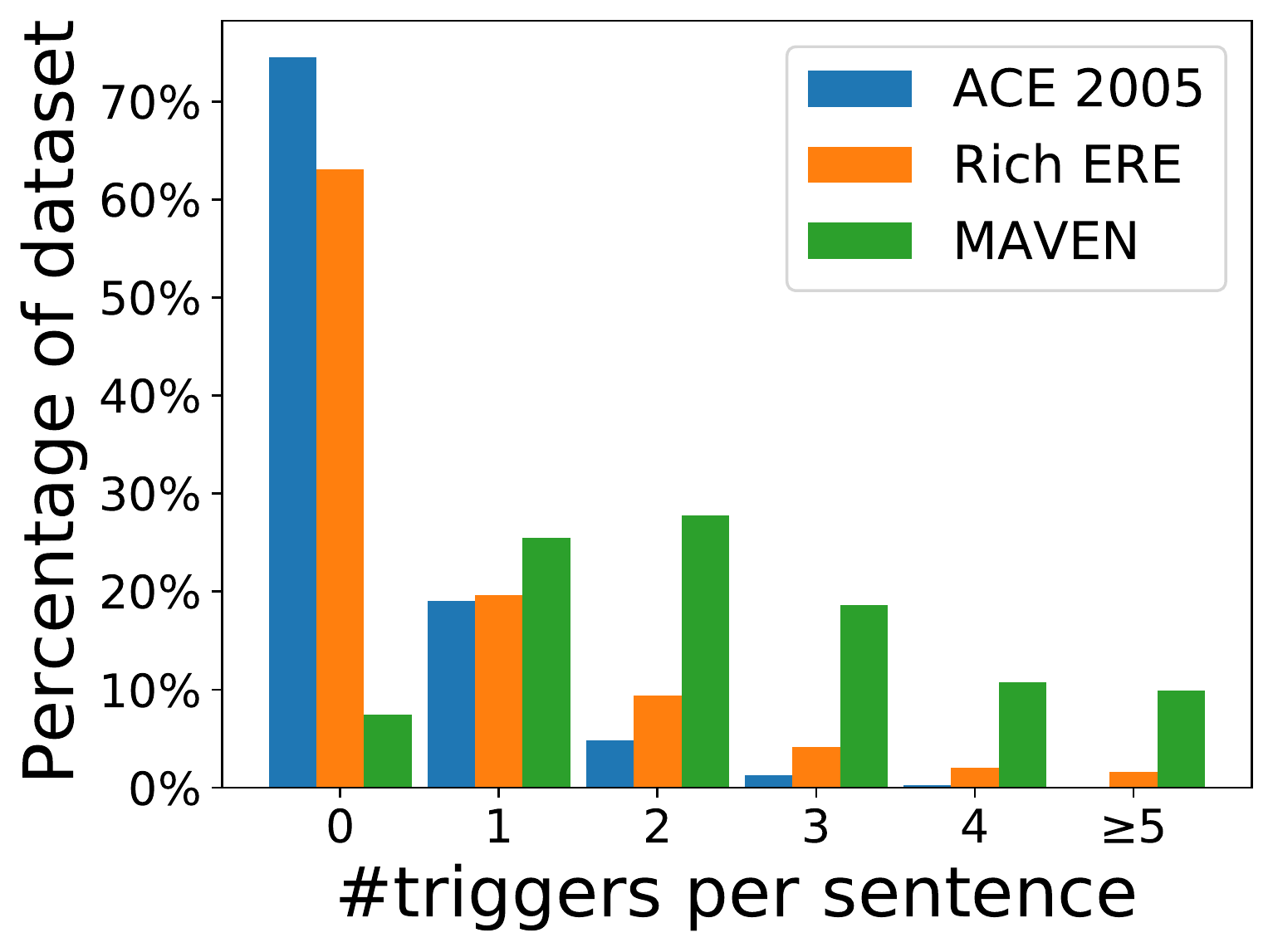}
}

\caption{Distribution of sentences containing different numbers of (golden) triggers of three datasets.}
\label{fig:multipleevent}
\end{figure}

\subsection{Data Size}

We show the main statistics of MAVEN and compare them with some existing widely-used ED datasets in Table~\ref{tab:dataset_statistics}, including the most widely-used ACE 2005 dataset~\cite{walker2006ace} and a series of Rich ERE annotation datasets provided by TAC KBP competition, which are DEFT Rich ERE English Training Annotation V2 (LDC2015E29), DEFT Rich ERE English Training Annotation R2 V2 (LDC2015E68), DEFT Rich ERE Chinese and English Parallel Annotation V2 (LDC2015E78), TAC KBP Event Nugget Data 2014-2016 (LDC2017E02)~\cite{ellis2014overview,ellis2015overview,ellis2016overview} and TAC KBP 2017 (LDC2017E55)~\cite{getman2017overview}. The Rich ERE datasets can be combined as used in \newcite{lin-etal-2019-cost} and \newcite{lu-etal-2019-distilling}, but the combined dataset is still much smaller than MAVEN. MAVEN is larger than all existing ED datasets, especially in the number of events. Hopefully, the large-scale dataset can accelerate the research on general domain ED.

\subsection{Data Distribution}

Figure~\ref{fig:ourDist} shows the histogram of MAVEN event types by their instance numbers. We can observe that the inherent data imbalance problem also exists in MAVEN. However, as MAVEN is large-scale, $41\%$ and $82\%$ event types have more than $500$ and $100$ instances respectively. Compared with existing datasets like ACE 2005 (only $39\%$ event types have more than 100 instances), MAVEN significantly alleviates the data scarcity problem, which will benefit developing strong ED models and various event-related downstream applications. 

We want MAVEN to serve as a real-world ED dataset, and the distribution of real-world data is inherently long-tail. To evaluate the ED ability on the long-tail scenario is also our goal. Hence, we do not apply data augmentation or balancing during dataset construction and maintain the real-world distribution in MAVEN. To support future exploration of handling the long-tail problem, we design a hierarchical event type schema, which may help transfer knowledge from the coarse-grained event types to the long-tail fine-grained types. We show the five top-level (most coarse-grained) types and their proportions in Table~\ref{tab:majortype} and the detailed hierarchical schema in Appendix~\ref{app:hierschema}.


\subsection{Multiple Events in One Sentence}

\label{sec:multipleevents}

A key phenomenon of ED datasets is that a sentence can express multiple events at the same time, and ED models will better classify the event types with the help of correlations between multiple events. Although the multiple event phenomenon has been investigated by existing works~\cite{li2013joint,chen-etal-2018-collective,liu-etal-2018-jointly} on ACE 2005 dataset, we observe that this phenomenon is much more common and complex on MAVEN.

In Figure~\ref{fig:multipleevent}, we compare MAVEN's percentages of sentences containing different numbers of triggers with ACE 2005 and the combined Rich ERE dataset (corresponding to the ``Total'' row in Table~\ref{tab:dataset_statistics}). We can observe that because MAVEN's coverage on general domain events is much higher, the multiple events in one sentence phenomenon is much more common in MAVEN than existing datasets. Moreover, as more event types are defined in MAVEN, the association relations between event types will be much more complex than on ACE 2005. We hope MAVEN can facilitate ED research on modeling multiple event correlations. 
\begin{table}[t]
    \centering
\scalebox{0.792}{
\begin{tabular}{lrrrr}
\toprule
\textbf{Subset} & \textbf{\#Document} & \textbf{\#Event}& \textbf{\#Mention}& \textbf{\#Negative.} \\ 
\midrule
Train        & $2,913$ & $73,496$         & $77,993$                 & $323,992$         \\
Dev      & $710$    & $17,726$          & $18,904$                 & $79,699$              \\
Test            & $857$     & $20,389$         & $21,835$      & $93,570$  \\
\bottomrule
\end{tabular}
}
    \caption{The statistics of splitting MAVEN. ``\#Negative.'' is the number of negative instances. }
    \label{tab:split_statistics}
\end{table}

\begin{table*}[t]
\centering
\scalebox{0.88}{
\begin{tabular}{l|ccc|ccc}
\toprule
\multirow{2}{*}{\textbf{Method}} & \multicolumn{3}{c|}{\textbf{ACE 2005}} & \multicolumn{3}{c}{\textbf{MAVEN}}        \\ \cmidrule{2-7} 
                        & \textbf{P}        & \textbf{R}        & \textbf{F-1}     & \textbf{P}         & \textbf{R}         & \textbf{F-1}       \\ \midrule
DMCNN                   & $73.7\pm2.42$     & $63.3\pm3.30$     & $68.0\pm1.95$    & $\bm{66.3\pm0.89}$ & $55.9\pm0.50$ & $60.6\pm0.20$ \\ 
BiLSTM                  & $71.7\pm1.70$     & $\bm{82.8\pm1.00}$     & $\bm{76.8\pm1.01}$    & $59.8\pm0.81$ & $67.0\pm0.76$ & $62.8\pm0.82$ \\ 
BiLSTM+CRF              & $\bm{77.2\pm2.08}$     & $74.9\pm2.62$     & $75.4\pm1.64$    & $63.4\pm0.70$ & $64.8\pm0.69$ & $64.1\pm0.13$ \\
MOGANED                 & $70.4\pm1.38$     & $73.9\pm2.24$     & $72.1\pm0.39$    & $63.4\pm0.88$ & $64.1\pm0.90$ & $63.8\pm0.18$ \\ 
DMBERT                  & $70.2\pm1.71$    & $78.9\pm1.64$    &  $74.3\pm0.81$     & $62.7\pm1.01$ & $\bm{72.3\pm1.03}$ & $67.1\pm0.41$ \\
BERT+CRF            & $71.3\pm1.77$    & $77.1\pm1.99$    &  $74.1\pm1.56$     & $65.0\pm0.84$ & $70.9\pm0.94$ & $\bm{67.8\pm0.15}$ \\
\bottomrule
\end{tabular}
}
\caption{The overall trigger classification performance of various models on ACE 2005 and MAVEN.}
\label{tab:mainResult}
\end{table*}

\section{Experiments}

Our experiments and analyses will show the challenges of MAVEN and promising ED directions.


\subsection{Benchmark Setting}

\label{sec:benchmarkSetting}
We firstly introduce the MAVEN benchmark setting here. MAVEN is randomly split into training, development, and test sets and the statistics of the three sets are shown in Table~\ref{tab:split_statistics}. After splitting, there are $32\%$ and $71\%$ of event types that have more than $500$ and $100$ training instances respectively, which ensures the models can be well-trained. 

Conventionally, the existing ED datasets only provide the standard annotation of positive instances (the annotated event triggers) and researchers will sample the negative instances (non-trigger words or phrases) by themselves, which may lead to potential unfair comparisons between different methods. In MAVEN, we provide official negative instances to ensure fair comparisons. As described in Section~\ref{sec:candidate}, the negative instances are the content words labeled by the NLTK POS tagger or the phrases which can be matched with the FrameNet lexical units. In other words, we only filter out those empty words, which will not influence the application of models developed on MAVEN.

\subsection{Experimental Setting}

\paragraph{Models}
Recently, various neural models have been developed for ED and achieved superior performances compared with traditional feature-based models. Hence, we reproduce six representative state-of-the-art neural models and report their performances on both MAVEN and widely-used ACE 2005 to assess the challenges of MAVEN, including:
(1) \textbf{DMCNN}~\cite{chen2015event} is a convolutional neural network (CNN) model, which leverages a CNN to automatically learn sequence representations and a dynamic multi-pooling mechanism to aggregate learned features into trigger-specific representations for classification.
(2) \textbf{BiLSTM}~\cite{hochreiter1997long} is a vanilla recurrent neural network baseline, which adopts the widely-used bi-directional long short-term memory network to learn textual representations, and then uses the hidden states at the positions of trigger candidates for classifying event types.
(3) \textbf{MOGANED}~\cite{yan-etal-2019-event} is an advanced graph neural network (GNN) model. It proposes a multi-order graph attention network to effectively model the multi-order syntactic relations in dependency trees and improve ED.
(4) \textbf{DMBERT}~\cite{wang-etal-2019-adversarial-training} is a vanilla BERT-based model. It takes advantage of the effective pre-trained language representation model BERT~\cite{devlin-etal-2019-bert} and also adopts the dynamic multi-pooling mechanism to aggregate features for ED. We use the BERT$_{\small \texttt{BASE}}$ architecture in our experiments.
(5) Different from the above token-level classification models, \textbf{BiLSTM+CRF} and \textbf{BERT+CRF} are sequence labeling models. To verify the effectiveness of modeling multiple event correlations, the two models both adopt the conditional random field (CRF)~\cite{lafferty2001conditional} as their output layers, which can model structured output dependencies. And they use BiLSTM and BERT$_{\small \texttt{BASE}}$ as their feature extractors respectively. 

As we manually tune hyperparameters and some training details, the results of reproduced models may be different from the original papers. Please refer to Appendix~\ref{app:hyper} for reproduction details.

\paragraph{Evaluation}
Following the widely-used setting introduced in Section~\ref{sec:taskDef}, we report the micro precision, recall, and F-1 scores for trigger classification as our evaluation metrics. For direct comparisons with the token-level classification models, we use span-based metrics for the sequence labeling baselines. 
On ACE 2005\footnote{\url{catalog.ldc.upenn.edu/LDC2006T06}}, we use $40$ newswire articles for test, $30$ random documents for development, and $529$ documents for training following previous work~\cite{chen2015event,wang-etal-2019-hmeae}, and sample all the unlabeled words as negative instances. To get stable results, we run each model $10$ times on both datasets and report the averages and standard deviations for each metric.

\begin{figure}[t]
\centering
\scalebox{0.8}{
\includegraphics[width = 0.45\textwidth]{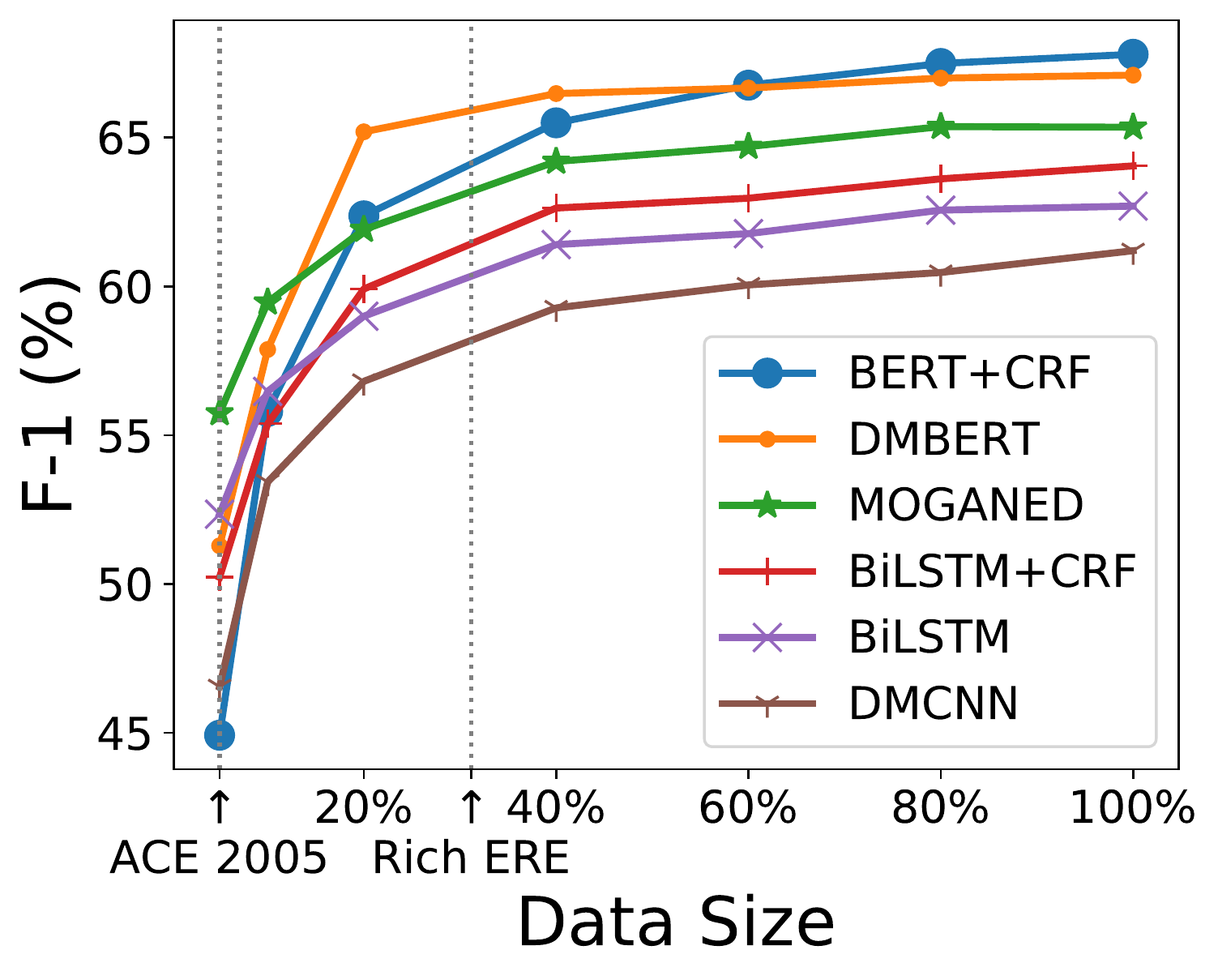}
}

\caption{Model performance (F-1) change along with the training data size.}
\label{fig:dataSize}
\end{figure}

\subsection{Overall Experimental Results}
\label{sec:overallResults}
The overall experimental results are in Table~\ref{tab:mainResult}, from which we have the following observations:

(1) Although the models perform well on ACE 2005, their performances are significantly lower and not satisfying on MAVEN. It indicates that our MAVEN is challenging and the general domain ED still needs more research efforts.
(2) The result deviations of various models on MAVEN are typically significantly lower than on the small-scale ACE 2005, which suggests that the small-scale datasets cannot stably benchmark sophisticated ED methods, while MAVEN alleviates this problem with its massive annotated data.
(3) It is surprising to find that the BiLSTM-based models achieve remarkably high performance on ACE 2005, even outperform the BERT models. We guess this is because the small-scale dataset cannot stably train and benchmark large models. The results on MAVEN are intuitive. 
(4) From the comparison between the \textbf{BiLSTM+CRF} and \textbf{BiLSTM}, we can observe that the CRF-based method achieves obvious improvement on MAVEN, but cannot outperform the vanilla \textbf{BiLSTM} on ACE 2005. \textbf{BERT+CRF} also outperforms \textbf{DMBERT} on MAVEN even without the effective dynamic multi-pooling mechanism. Considering the key advantage of the CRF output layer in ED is to model multiple event correlations, the results are consistent with our observations in Section~\ref{sec:multipleevents} that the multiple events in one sentence phenomenon is much more common in MAVEN. This suggests how to better modeling multiple events is worth exploring. 

\begin{figure}[t]
\centering
\scalebox{0.8}{

\includegraphics[width = 0.45\textwidth]{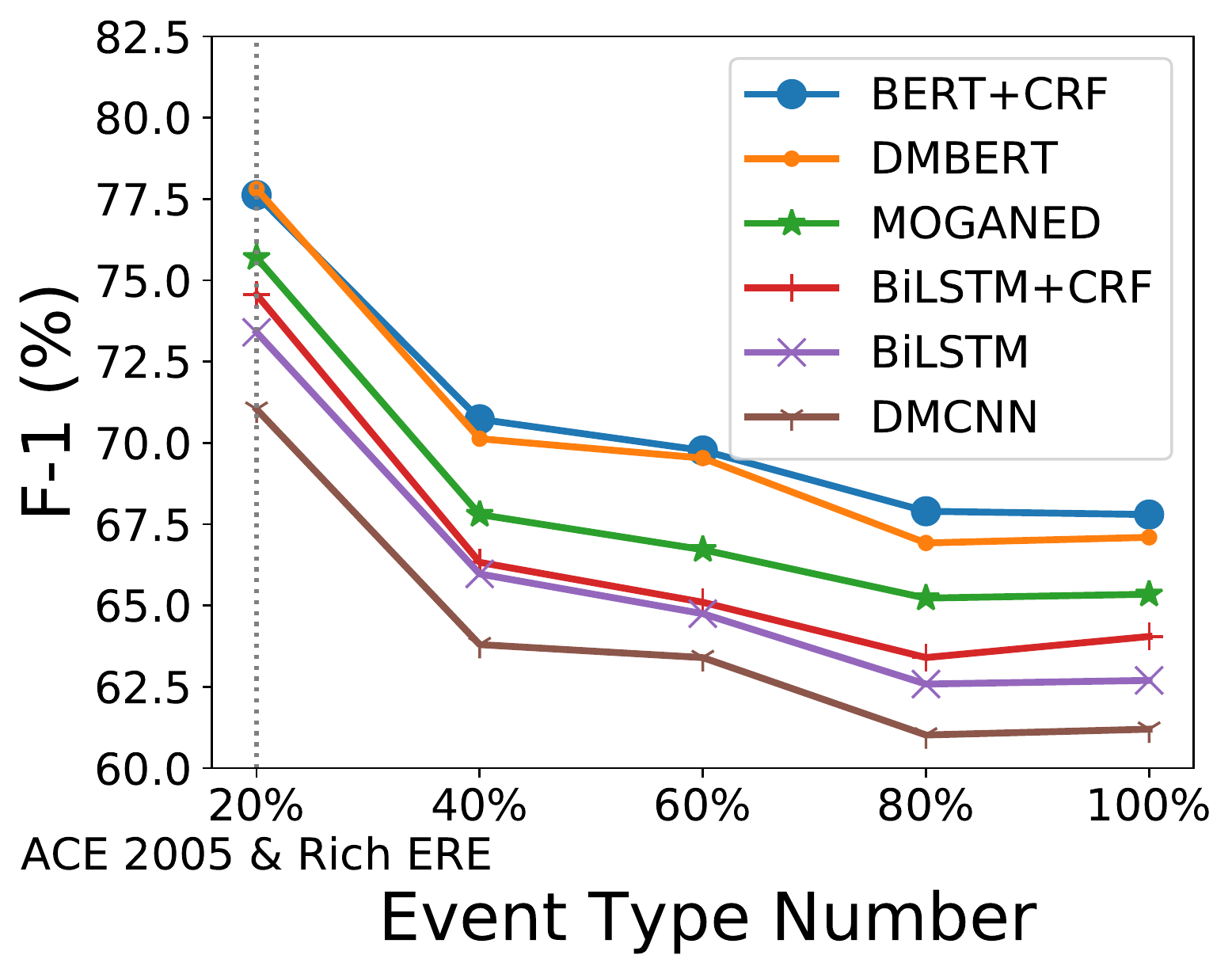}
}

\caption{Model performance (F-1) change along with the number of event types.}
\label{fig:typeNmber}
\end{figure}

\subsection{Analyses on Data Size and \#Event Types}

MAVEN contains more data and covers more event types compared with existing benchmarks. In this section, we analyze the benefits of a larger data scale and the challenge of more event types.

We randomly choose different proportions of documents from the MAVEN training set and compare the model performances trained with different sizes of data in Figure~\ref{fig:dataSize}. We can observe that MAVEN can sufficiently train the models and stably benchmark them, and we will get unreliable comparison results at the existing datasets' scale. 

We also randomly choose different proportions of event types and train the models to only classify the chosen types. The model performances are shown in Figure~\ref{fig:typeNmber}. With the increase in the number of event types, we can observe significant performance drops, which demonstrates the challenge brought by the high coverage of MAVEN. 

\begin{table}[t]
\centering
\scalebox{0.77}{
\begin{tabular}{l|ccc}
\toprule
\multirow{2}{*}{\textbf{Method}} & \multicolumn{3}{c}{\textbf{ACE 2005 Trigger Classification}}               \\ \cmidrule{2-4} 
                        & \textbf{P}             &\textbf{R}             & \textbf{F-1}                               \\ \midrule
DMBERT                  & $70.2\pm1.71$ & $\bm{78.9\pm1.64}$ & \multicolumn{1}{l}{$74.3\pm0.81$} \\ \midrule
\quad+aug                    & $68.7\pm1.21$       & $76.4\pm1.16$       & $72.4\pm0.75$                           \\
\quad+pretrain               & $\bm{71.9\pm1.12}$  & $78.7\pm1.44$ & $\bm{75.1\pm0.56}$                     \\ \bottomrule
\end{tabular}
}
\caption{The performance of DMBERT with two simple knowledge transfer methods on ACE 2005.}
\label{tab:trans}
\end{table}

\subsection{Analyses on Transferability}
\label{sec:transfer}
As MAVEN annotates a large range of general domain events, an intuitive question is whether the general ED knowledge learned on MAVEN can transfer to other ED tasks that do not have sufficient data. We examine the transferability of MAVEN with experiments on ACE 2005.

We explore two simple transfer learning methods on \textbf{DMBERT} model. (1) Data augmentation (\textbf{+aug}) is to add $18,729$ MAVEN instances into ACE 2005 training set and directly train the model. As the event schema of ACE 2005 and MAVEN is different, we manually build an incomplete mapping of event types, which is shown in Appendix~\ref{app:acemapping}. (2) Intermediate pre-training (\textbf{+pretrain}), which is to first train the model on MAVEN and then fine-tune it on ACE 2005. This method has been shown to be effective on some natural language inference tasks~\cite{wang-etal-2019-tell}. 

The results are shown in Table~\ref{tab:trans}, from which we can observe that as MAVEN focuses on different event types and a different text domain (Wikipedia), direct data augmentation harms ED performances while tested on ACE 2005 (newswire data). However, intermediate pre-training can improve ED on ACE 2005 with the general event knowledge learned on MAVEN, which indicates MAVEN's high coverage of event types can benefit other ED tasks. It is worth to explore how to apply more advanced transfer learning methods to improve the performance on low-resource ED scenarios.

\subsection{Error Analysis}
\label{sec:errorAna}
\begin{table}[t]
\centering
\small
\scalebox{0.8}{
\begin{tabular}{l|cc|ccc}
\toprule
\multicolumn{1}{c|}{\multirow{2}{*}{\textbf{Method}}} & \multicolumn{2}{c|}{\begin{tabular}[c]{@{}c@{}}\textbf{Identification} \\ \textbf{Mistakes}\end{tabular}}                             & \multicolumn{3}{c}{\textbf{Event Type Mistakes}}                                                                                                                                                  \\ \cmidrule{2-6} 
\multicolumn{1}{c|}{}                        & \begin{tabular}[c]{@{}c@{}}\textbf{FP}\end{tabular} & \begin{tabular}[c]{@{}c@{}}\textbf{FN}\end{tabular} & \begin{tabular}[c]{@{}c@{}}\textbf{Parent}\\ \textbf{-Children}\end{tabular} & \begin{tabular}[c]{@{}c@{}}\textbf{Between} \\ \textbf{Siblings}\end{tabular} & \begin{tabular}[c]{@{}c@{}}\textbf{Into} \\ \textbf{Top 50\%}\end{tabular} \\ \midrule
DMCNN                                        & $27.3\%$                                                 & $55.9\%$                                                 & $15.5\%$                                                   & $19.8\%$                                                    & $89.2\%$                                                       \\
BiLSTM                                       & $26.9\%$                                                 & $52.9\%$                                                 & $14.5\%$                                                   & $14.6\%$                                                    & $90.3\%$                                                       \\
MOGANED                                      & $44.5\%$                                                 & $31.3\%$                                                 & $15.5\%$                                                   & $17.8\%$                                                    & $86.8\%$                                                       \\
DMBERT                                       & $48.5\%$                                                 & $27.2\%$                                                 & $13.1\%$                                                   & $19.0\%$                                                    & $87.0\%$                                                       \\ \bottomrule
\end{tabular}
}
\caption{The proportions of different kinds of mistakes in various models' predictions on MAVEN dev set. The numbers of positive and negative instances are $18,904$ and $79,699$, respectively.}
\label{tab:error}
\end{table}

To analyze the abilities required by MAVEN, we conduct error analyses on the prediction results of various token-level classification ED models (the sequence labeling methods have span prediction errors, hence cannot be analyzed with misclassifying types as here). The results are shown in Table~\ref{tab:error}, from which we can observe:

(1) ``Identification Mistakes'' indicates misclassifying negative instances into positive types (FP) or misclassifying positive instances into \texttt{N/A} (FN), which is the most common mistake. It indicates that identifying event semantics from various and complicated language expressions is still challenging and needs further efforts.

(2) ``Event Type Mistakes'' indicates misclassifying between the $168$ event types. The percentages of the three subtype mistakes are all calculated within ``Event Type Mistakes''. ``Parent-Children'' indicates misclassifying instances into their parent or children types in the tree-structure hierarchical event type schema, and ``Between Siblings'' indicates misclassifying instances into their sibling types. Considering each event type only has one parent type and $9.96$ sibling types on average, the percentages of these two kinds of mistakes are significantly higher than misclassifying into other distant types. It suggests that existing models typically cannot well distinguish subtle differences between event types, and our hierarchical event type schema may help models to this point.

(3) ``Into Top $50\%$'' indicates misclassifying into event types with top $50\%$ amounts of data. It shows that ED models should develop the ability to resist the influence of the inherent data imbalance problem. Hence, further explorations on handling these problems may bring more effective ED models. To this end, our hierarchical event schema may also be helpful in developing data balancing and data augmentation methods.


\section{Related Work}
As stated in Section~\ref{sec:taskDef}, we follow the ED task definition specified in the ACE challenges, especially the ACE 2005 dataset~\cite{doddington2004automatic} in this paper, which requires ED models to generally detect the event triggers and classify them into specific event types. The ACE event schema is simplified into Light ERE and further extended to Rich ERE~\cite{song-etal-2015-light} to cover more but still a limited number of event types. Rich ERE is used to create various datasets and the TAC KBP challenges~\cite{ellis2014overview,ellis2015overview,ellis2016overview,getman2017overview}. Nowadays, the majority of ED and event extraction models~\cite{ji2008refining,li2013joint,chen2015event,feng2016language,liu-etal-2017-exploiting,zhao-etal-2018-document,yan-etal-2019-event} are developed on these datasets. Our MAVEN follows the effective framework and extends it to numerous general domain event types and data instances.

There are also various datasets defining the ED task in different ways. The early MUC series datasets~\cite{grishman-sundheim-1996-message} define event extraction as a slot-filling task. The TDT corpus~\cite{allan2012topic} and some recent datasets~\cite{minard-etal-2016-meantime,araki-mitamura-2018-open,sims-etal-2019-literary,liu-etal-2019-open} follow the open-domain paradigm, which does not require models to classify events into pre-defined event types for better coverage but limits the downstream application of the extracted events. Some datasets are developed for ED on specific domains, like the bio-medical domain~\cite{pyysalo2007bioinfer,kim2008text,thompson2009construction,buyko2010genereg,nedellec2013overview}, literature~\cite{sims-etal-2019-literary}, Twitter~\cite{ritter2012open,guo-etal-2013-linking} and breaking news~\cite{pustejovsky2003timebank}. These datasets are also typically small-scale due to the inherent complexity of event annotation, but their different settings are complementary to our work.

\section{Conclusion and Future work}
In this paper, we present a massive general domain event detection dataset (MAVEN), which significantly alleviates the data scarcity and low coverage problems of existing datasets. We conduct a thorough evaluation of the state-of-the-art ED models on MAVEN. The results indicate that general domain ED is still challenging and MAVEN may facilitate further research. We also explore some promising directions with analytic experiments, including modeling multiple event correlations (Section~\ref{sec:overallResults}), utilizing the hierarchical event schema to distinguish close types (Section~\ref{sec:errorAna}), and improving other ED tasks with transfer learning (Section~\ref{sec:transfer}). In the future, we will extend MAVEN to more event-related tasks like event argument extraction, event sequencing, etc.

\section*{Acknowledgement}
We thank the anonymous reviewers for their insightful comments and suggestions. This work is supported by NSFC Key Projects (U1736204, 61533018), grants from Institute for Guo Qiang, Tsinghua University (2019GQB0003) and Beijing Academy of Artificial Intelligence (BAAI2019ZD0502). This work is also supported by the Pattern Recognition Center, WeChat AI, Tencent Inc. Xiaozhi Wang is supported by Tsinghua University Initiative Scientific Research Program.

\bibliography{emnlp2020}

\begin{thebibliography}{73}
\expandafter\ifx\csname natexlab\endcsname\relax\def\natexlab#1{#1}\fi

\bibitem[{Aguilar et~al.(2014)Aguilar, Beller, McNamee, Van~Durme, Strassel,
  Song, and Ellis}]{aguilar-etal-2014-comparison}
Jacqueline Aguilar, Charley Beller, Paul McNamee, Benjamin Van~Durme, Stephanie
  Strassel, Zhiyi Song, and Joe Ellis. 2014.
\newblock \href {https://doi.org/10.3115/v1/W14-2907} {A comparison of the
  events and relations across {ACE}, {ERE}, {TAC}-{KBP}, and {F}rame{N}et
  annotation standards}.
\newblock In \emph{Proceedings of the Second Workshop on {EVENTS}: Definition,
  Detection, Coreference, and Representation}, pages 45--53.

\bibitem[{Ahn(2006)}]{ahn2006stages}
David Ahn. 2006.
\newblock \href {http://aclweb.org/anthology/W06-0901} {The stages of event
  extraction}.
\newblock In \emph{Proceedings of ACL Workshop on Annotating and Reasoning
  about Time and Events}, pages 1--8.

\bibitem[{Allan(2012)}]{allan2012topic}
James Allan. 2012.
\newblock \href {https://www.springer.com/gp/book/9780792376644} {\emph{Topic
  detection and tracking: event-based information organization}}, volume~12.
\newblock Springer Science \& Business Media.

\bibitem[{Araki and Mitamura(2015)}]{araki2015joint}
Jun Araki and Teruko Mitamura. 2015.
\newblock \href {https://doi.org/10.18653/v1/D15-1247} {Joint event trigger
  identification and event coreference resolution with structured perceptron}.
\newblock In \emph{Proceedings of EMNLP}, pages 2074--2080.

\bibitem[{Araki and Mitamura(2018)}]{araki-mitamura-2018-open}
Jun Araki and Teruko Mitamura. 2018.
\newblock \href {https://www.aclweb.org/anthology/C18-1075} {Open-domain event
  detection using distant supervision}.
\newblock In \emph{Proceedings of COLING}, pages 878--891.

\bibitem[{Baker et~al.(1998)Baker, Fillmore, and Lowe}]{baker1998berkeley}
Collin~F. Baker, Charles~J. Fillmore, and John~B. Lowe. 1998.
\newblock \href {https://doi.org/10.3115/980845.980860} {The {B}erkeley
  {F}rame{N}et project}.
\newblock In \emph{Proceedings of ACL-COLING}, pages 86--90.

\bibitem[{Basile et~al.(2014)Basile, Caputo, Semeraro, and
  Siciliani}]{dart2014wsBasile}
P~Basile, A~Caputo, G~Semeraro, and L~Siciliani. 2014.
\newblock \href {http://ceur-ws.org/Vol-1314/paper-04.pdf} {Extending an
  information retrieval system through time event extraction}.
\newblock In \emph{Proceedings of DART}, pages 36--47.

\bibitem[{Bird(2006)}]{bird-2006-nltk}
Steven Bird. 2006.
\newblock \href {https://doi.org/10.3115/1225403.1225421} {{NLTK}: The
  {N}atural {L}anguage {T}oolkit}.
\newblock In \emph{Proceedings of the {COLING}/{ACL} 2006 Interactive
  Presentation Sessions}, pages 69--72.

\bibitem[{Buyko et~al.(2010)Buyko, Beisswanger, and Hahn}]{buyko2010genereg}
Ekaterina Buyko, Elena Beisswanger, and Udo Hahn. 2010.
\newblock \href
  {http://www.lrec-conf.org/proceedings/lrec2010/pdf/407_Paper.pdf} {The
  {G}ene{R}eg corpus for gene expression regulation events {---} an overview of
  the corpus and its in-domain and out-of-domain interoperability}.
\newblock In \emph{Proceedings of LREC}.

\bibitem[{Chen et~al.(2017)Chen, Liu, Zhang, Liu, and
  Zhao}]{chen-etal-2017-automatically}
Yubo Chen, Shulin Liu, Xiang Zhang, Kang Liu, and Jun Zhao. 2017.
\newblock \href {https://doi.org/10.18653/v1/P17-1038} {Automatically labeled
  data generation for large scale event extraction}.
\newblock In \emph{Proceedings of ACL}, pages 409--419.

\bibitem[{Chen et~al.(2015)Chen, Xu, Liu, Zeng, and Zhao}]{chen2015event}
Yubo Chen, Liheng Xu, Kang Liu, Daojian Zeng, and Jun Zhao. 2015.
\newblock \href {https://doi.org/10.3115/v1/P15-1017} {Event extraction via
  dynamic multi-pooling convolutional neural networks}.
\newblock In \emph{Proceedings of ACL-IJCNLP}, pages 167--176.

\bibitem[{Chen et~al.(2018)Chen, Yang, Liu, Zhao, and
  Jia}]{chen-etal-2018-collective}
Yubo Chen, Hang Yang, Kang Liu, Jun Zhao, and Yantao Jia. 2018.
\newblock \href {https://doi.org/10.18653/v1/D18-1158} {Collective event
  detection via a hierarchical and bias tagging networks with gated multi-level
  attention mechanisms}.
\newblock In \emph{Proceedings of EMNLP}, pages 1267--1276.

\bibitem[{Cheng and Erk(2018)}]{cheng2018implicit}
Pengxiang Cheng and Katrin Erk. 2018.
\newblock \href {https://doi.org/10.18653/v1/N18-1076} {Implicit argument
  prediction with event knowledge}.
\newblock In \emph{Proceedings of ACL}, pages 831--840.

\bibitem[{Cohen(1960)}]{cohen1960coefficient}
Jacob Cohen. 1960.
\newblock \href {https://doi.org/10.1177/001316446002000104} {A coefficient of
  agreement for nominal scales}.
\newblock \emph{Educational and psychological measurement}, 20(1):37--46.

\bibitem[{Consortium(2005)}]{linguistic2005ace}
Linguistic~Data Consortium. 2005.
\newblock \href
  {https://www.ldc.upenn.edu/sites/www.ldc.upenn.edu/files/english-events-guidelines-v5.4.3.pdf}
  {{ACE} ({A}utomatic {C}ontent {E}xtraction) {E}nglish annotation guidelines
  for events}.
\newblock \emph{Version}, 5(4).

\bibitem[{Devlin et~al.(2019)Devlin, Chang, Lee, and
  Toutanova}]{devlin-etal-2019-bert}
Jacob Devlin, Ming-Wei Chang, Kenton Lee, and Kristina Toutanova. 2019.
\newblock \href {https://doi.org/10.18653/v1/N19-1423} {{BERT}: Pre-training of
  deep bidirectional transformers for language understanding}.
\newblock In \emph{Proceedings of NAACL-HLT}, pages 4171--4186.

\bibitem[{Ding et~al.(2019)Ding, Li, Liu, Zheng, and
  Lin}]{ding-etal-2019-event}
Ning Ding, Ziran Li, Zhiyuan Liu, Haitao Zheng, and Zibo Lin. 2019.
\newblock \href {https://doi.org/10.18653/v1/D19-1033} {Event detection with
  trigger-aware lattice neural network}.
\newblock In \emph{Proceedings of EMNLP-IJCNLP}, pages 347--356.

\bibitem[{Doddington et~al.(2004)Doddington, Mitchell, Przybocki, Ramshaw,
  Strassel, and Weischedel}]{doddington2004automatic}
George Doddington, Alexis Mitchell, Mark Przybocki, Lance Ramshaw, Stephanie
  Strassel, and Ralph Weischedel. 2004.
\newblock \href {http://www.lrec-conf.org/proceedings/lrec2004/pdf/5.pdf} {The
  automatic content extraction ({ACE}) program {--} tasks, data, and
  evaluation}.
\newblock In \emph{Proceedings of {LREC}}.

\bibitem[{Dong and Dong(2003)}]{dong2003hownet}
Zhendong Dong and Qiang Dong. 2003.
\newblock \href {https://ieeexplore.ieee.org/document/1276017} {{HowNet} - a
  hybrid language and knowledge resource}.
\newblock In \emph{Proceedings of NLP-KE}, pages 820--824.

\bibitem[{Ellis et~al.(2015)Ellis, Getman, Fore, Kuster, Song, Bies, and
  Strassel}]{ellis2015overview}
Joe Ellis, Jeremy Getman, Dana Fore, Neil Kuster, Zhiyi Song, Ann Bies, and
  Stephanie~M Strassel. 2015.
\newblock \href
  {https://www.ldc.upenn.edu/sites/www.ldc.upenn.edu/files/tackbp2015_overview.pdf}
  {Overview of linguistic resources for the {TAC} {KBP} 2015 evaluations:
  {M}ethodologies and results.}
\newblock In \emph{TAC}.

\bibitem[{Ellis et~al.(2016)Ellis, Getman, Fore, Kuster, Song, Bies, and
  Strassel}]{ellis2016overview}
Joe Ellis, Jeremy Getman, Dana Fore, Neil Kuster, Zhiyi Song, Ann Bies, and
  Stephanie~M Strassel. 2016.
\newblock \href
  {https://www.ldc.upenn.edu/sites/www.ldc.upenn.edu/files/tackbp2016-linguistic-resources-tackbp-1.pdf}
  {Overview of linguistic resources for the {TAC} {KBP} 2016 evaluations:
  {M}ethodologies and results.}
\newblock In \emph{TAC}.

\bibitem[{Ellis et~al.(2014)Ellis, Getman, and Strassel}]{ellis2014overview}
Joe Ellis, Jeremy Getman, and Stephanie~M Strassel. 2014.
\newblock \href
  {https://www.ldc.upenn.edu/sites/www.ldc.upenn.edu/files/tackbp-2014-overview.pdf}
  {Overview of linguistic resources for the {TAC} {KBP} 2014 evaluations:
  {P}lanning, execution, and results}.
\newblock In \emph{TAC}.

\bibitem[{Feng et~al.(2016)Feng, Huang, Tang, Qin, Ji, and
  Liu}]{feng2016language}
Xiaocheng Feng, Lifu Huang, Duyu Tang, Bing Qin, Heng Ji, and Ting Liu. 2016.
\newblock \href {https://doi.org/10.18653/v1/P16-2011} {A language-independent
  neural network for event detection}.
\newblock In \emph{Proceedings of ACL}, pages 66--71.

\bibitem[{Ferguson et~al.(2018)Ferguson, Lockard, Weld, and
  Hajishirzi}]{ferguson-etal-2018-semi}
James Ferguson, Colin Lockard, Daniel Weld, and Hannaneh Hajishirzi. 2018.
\newblock \href {https://doi.org/10.18653/v1/N18-2058} {Semi-supervised event
  extraction with paraphrase clusters}.
\newblock In \emph{Proceedings of NAACL}, pages 359--364.

\bibitem[{Fillmore(2006)}]{fillmore2006frame}
Charles Fillmore. 2006.
\newblock Frame semantics.
\newblock \emph{Cognitive linguistics: Basic readings}, 34:373--400.

\bibitem[{Fillmore(1976)}]{fillmore1976frame}
Charles~J Fillmore. 1976.
\newblock \href {http://www.icsi.berkeley.edu/pubs/ai/framesemantics76.pdf}
  {Frame semantics and the nature of language}.
\newblock In \emph{Annals of the New York Academy of Sciences: Conference on
  the origin and development of language and speech}, volume 280, pages 20--32.

\bibitem[{Ge et~al.(2018)Ge, Cui, Chang, Sui, Wei, and
  Zhou}]{ge-etal-2018-eventwiki}
Tao Ge, Lei Cui, Baobao Chang, Zhifang Sui, Furu Wei, and Ming Zhou. 2018.
\newblock \href {https://www.aclweb.org/anthology/L18-1079} {{E}vent{W}iki: A
  knowledge base of major events}.
\newblock In \emph{Proceedings of LREC}.

\bibitem[{Getman et~al.(2017)Getman, Ellis, Song, Tracey, and
  Strassel}]{getman2017overview}
Jeremy Getman, Joe Ellis, Zhiyi Song, Jennifer Tracey, and Stephanie~M
  Strassel. 2017.
\newblock \href
  {https://tac.nist.gov/publications/2017/additional.papers/TAC2017.KBP_resources_overview.proceedings.pdf}
  {Overview of linguistic resources for the {TAC} {KBP} 2017 evaluations:
  {M}ethodologies and results.}
\newblock In \emph{TAC}.

\bibitem[{Ghaeini et~al.(2016)Ghaeini, Fern, Huang, and
  Tadepalli}]{ghaeini2016event}
Reza Ghaeini, Xiaoli Fern, Liang Huang, and Prasad Tadepalli. 2016.
\newblock \href {https://doi.org/10.18653/v1/P16-2060} {Event nugget detection
  with forward-backward recurrent neural networks}.
\newblock In \emph{Proceedings of ACL}, pages 369--373.

\bibitem[{Grishman and Sundheim(1996)}]{grishman-sundheim-1996-message}
Ralph Grishman and Beth Sundheim. 1996.
\newblock \href {https://www.aclweb.org/anthology/C96-1079} {Message
  understanding conference- 6: A brief history}.
\newblock In \emph{Proceedings of {COLING}}.

\bibitem[{Guo et~al.(2013)Guo, Li, Ji, and Diab}]{guo-etal-2013-linking}
Weiwei Guo, Hao Li, Heng Ji, and Mona Diab. 2013.
\newblock \href {https://www.aclweb.org/anthology/P13-1024} {Linking tweets to
  news: A framework to enrich short text data in social media}.
\newblock In \emph{Proceedings of ACL}, pages 239--249.

\bibitem[{Gupta and Ji(2009)}]{gupta-ji:2009:Short}
Prashant Gupta and Heng Ji. 2009.
\newblock \href {http://www.aclweb.org/anthology/P/P09/P09-2093} {Predicting
  unknown time arguments based on cross-event propagation}.
\newblock In \emph{Proceedings of ACL-IJCNLP}, pages 369--372.

\bibitem[{Hochreiter and Schmidhuber(1997)}]{hochreiter1997long}
Sepp Hochreiter and J{\"u}rgen Schmidhuber. 1997.
\newblock \href {https://www.bioinf.jku.at/publications/older/2604.pdf} {Long
  short-term memory}.
\newblock \emph{Neural computation}, 9(8):1735--1780.

\bibitem[{Huang et~al.(2016)Huang, Cassidy, Feng, Ji, Voss, Han, and
  Sil}]{huang-etal-2016-liberal}
Lifu Huang, Taylor Cassidy, Xiaocheng Feng, Heng Ji, Clare~R. Voss, Jiawei Han,
  and Avirup Sil. 2016.
\newblock \href {https://doi.org/10.18653/v1/P16-1025} {Liberal event
  extraction and event schema induction}.
\newblock In \emph{Proceedings of ACL}, pages 258--268.

\bibitem[{Huang et~al.(2018)Huang, Ji, Cho, Dagan, Riedel, and
  Voss}]{huang-etal-2018-zero}
Lifu Huang, Heng Ji, Kyunghyun Cho, Ido Dagan, Sebastian Riedel, and Clare
  Voss. 2018.
\newblock \href {https://doi.org/10.18653/v1/P18-1201} {Zero-shot transfer
  learning for event extraction}.
\newblock In \emph{Proceedings of ACL}, pages 2160--2170.

\bibitem[{Ji and Grishman(2008)}]{ji2008refining}
Heng Ji and Ralph Grishman. 2008.
\newblock \href {http://aclweb.org/anthology/P08-1030} {Refining event
  extraction through cross-document inference}.
\newblock In \emph{Proceedings of ACL}, pages 254--262.

\bibitem[{Kim et~al.(2008)Kim, Ohta, Oda, and Tsujii}]{kim2008text}
Jin-Dong Kim, Tomoko Ohta, Kanae Oda, and Jun'ichi Tsujii. 2008.
\newblock \href {https://doi.org/10.1142/9781848161092_0019} {{From} {Text} to
  {Pathway}: {Corpus} annotation for knowledge acquisition from biomedical
  literature}.
\newblock \emph{Series on Advances in Bioinformatics and Computational
  Biology}, 6:165--176.

\bibitem[{Kingma and Ba(2014)}]{adam}
Diederik Kingma and Jimmy Ba. 2014.
\newblock \href {https://arxiv.org/abs/1412.6980} {Adam: A method for
  stochastic optimization}.
\newblock In \emph{Proceedings of ICLR}.

\bibitem[{Lafferty et~al.(2001)Lafferty, McCallum, and
  Pereira}]{lafferty2001conditional}
John~D. Lafferty, Andrew McCallum, and Fernando C.~N. Pereira. 2001.
\newblock \href {https://doi.org/10.5555/645530.655813} {Conditional random
  fields: Probabilistic models for segmenting and labeling sequence data}.
\newblock In \emph{Proceedings of ICML}, page 282–289.

\bibitem[{Li et~al.(2013)Li, Ji, and Huang}]{li2013joint}
Qi~Li, Heng Ji, and Liang Huang. 2013.
\newblock \href {http://aclweb.org/anthology/P13-1008} {Joint event extraction
  via structured prediction with global features}.
\newblock In \emph{Proceedings of ACL}, pages 73--82.

\bibitem[{Li et~al.(2019)Li, Cheng, He, Wang, and Jin}]{Li2019JointEE}
Wei Li, Dezhi Cheng, Lei He, Yuanzhuo Wang, and Xiaolong Jin. 2019.
\newblock \href {https://ieeexplore.ieee.org/document/8643786} {Joint event
  extraction based on hierarchical event schemas from framenet}.
\newblock \emph{IEEE Access}, 7:25001--25015.

\bibitem[{Lin et~al.(2019)Lin, Lu, Han, and Sun}]{lin-etal-2019-cost}
Hongyu Lin, Yaojie Lu, Xianpei Han, and Le~Sun. 2019.
\newblock \href {https://doi.org/10.18653/v1/P19-1521} {Cost-sensitive
  regularization for label confusion-aware event detection}.
\newblock In \emph{Proceedings of ACL}, pages 5278--5283.

\bibitem[{Liu et~al.(2017)Liu, Chen, Liu, and Zhao}]{liu-etal-2017-exploiting}
Shulin Liu, Yubo Chen, Kang Liu, and Jun Zhao. 2017.
\newblock \href {https://doi.org/10.18653/v1/P17-1164} {Exploiting argument
  information to improve event detection via supervised attention mechanisms}.
\newblock In \emph{Proceedings of ACL}, pages 1789--1798.

\bibitem[{Liu et~al.(2019)Liu, Huang, and Zhang}]{liu-etal-2019-open}
Xiao Liu, Heyan Huang, and Yue Zhang. 2019.
\newblock \href {https://doi.org/10.18653/v1/P19-1276} {Open domain event
  extraction using neural latent variable models}.
\newblock In \emph{Proceedings of ACL}, pages 2860--2871.

\bibitem[{Liu et~al.(2018)Liu, Luo, and Huang}]{liu-etal-2018-jointly}
Xiao Liu, Zhunchen Luo, and Heyan Huang. 2018.
\newblock \href {https://doi.org/10.18653/v1/D18-1156} {Jointly multiple events
  extraction via attention-based graph information aggregation}.
\newblock In \emph{Proceedings of EMNLP}, pages 1247--1256.

\bibitem[{Lu et~al.(2019)Lu, Lin, Han, and Sun}]{lu-etal-2019-distilling}
Yaojie Lu, Hongyu Lin, Xianpei Han, and Le~Sun. 2019.
\newblock \href {https://doi.org/10.18653/v1/P19-1429} {Distilling
  discrimination and generalization knowledge for event detection via
  delta-representation learning}.
\newblock In \emph{Proceedings of ACL}, pages 4366--4376.

\bibitem[{Minard et~al.(2016)Minard, Speranza, Urizar, Altuna, van Erp, Schoen,
  and van Son}]{minard-etal-2016-meantime}
Anne-Lyse Minard, Manuela Speranza, Ruben Urizar, Bego{\~n}a Altuna, Marieke
  van Erp, Anneleen Schoen, and Chantal van Son. 2016.
\newblock \href {https://www.aclweb.org/anthology/L16-1699} {{MEANTIME}, the
  {N}ews{R}eader multilingual event and time corpus}.
\newblock In \emph{Proceedings of {LREC}}, pages 4417--4422.

\bibitem[{Mintz et~al.(2009)Mintz, Bills, Snow, and
  Jurafsky}]{mintz-etal-2009-distant}
Mike Mintz, Steven Bills, Rion Snow, and Daniel Jurafsky. 2009.
\newblock \href {https://www.aclweb.org/anthology/P09-1113} {Distant
  supervision for relation extraction without labeled data}.
\newblock In \emph{Proceedings of ACL}, pages 1003--1011.

\bibitem[{N{\'e}dellec et~al.(2013)N{\'e}dellec, Bossy, Kim, Kim, Ohta,
  Pyysalo, and Zweigenbaum}]{nedellec2013overview}
Claire N{\'e}dellec, Robert Bossy, Jin-Dong Kim, Jung-jae Kim, Tomoko Ohta,
  Sampo Pyysalo, and Pierre Zweigenbaum. 2013.
\newblock \href {https://www.aclweb.org/anthology/W13-2001} {Overview of
  {B}io{NLP} shared task 2013}.
\newblock In \emph{Proceedings of the {B}io{NLP} Shared Task 2013 Workshop},
  pages 1--7.

\bibitem[{Nguyen et~al.(2016)Nguyen, Cho, and Grishman}]{nguyen2016joint}
Thien~Huu Nguyen, Kyunghyun Cho, and Ralph Grishman. 2016.
\newblock \href {https://doi.org/10.18653/v1/N16-1034} {Joint event extraction
  via recurrent neural networks}.
\newblock In \emph{Proceedings of NAACL}, pages 300--309.

\bibitem[{Nguyen and Grishman(2015)}]{nguyen-grishman-2015-event}
Thien~Huu Nguyen and Ralph Grishman. 2015.
\newblock \href {https://doi.org/10.3115/v1/P15-2060} {Event detection and
  domain adaptation with convolutional neural networks}.
\newblock In \emph{Proceedings of ACL}, pages 365--371.

\bibitem[{Pennington et~al.(2014)Pennington, Socher, and
  Manning}]{pennington-etal-2014-glove}
Jeffrey Pennington, Richard Socher, and Christopher Manning. 2014.
\newblock \href {https://doi.org/10.3115/v1/D14-1162} {{G}love: {G}lobal
  vectors for word representation}.
\newblock In \emph{Proceedings of EMNLP}, pages 1532--1543.

\bibitem[{Pustejovsky et~al.(2003)Pustejovsky, Hanks, Saurí, See, Gaizauskas,
  Setzer, Radev, Sundheim, Day, Ferro, and Lazo}]{pustejovsky2003timebank}
James Pustejovsky, Patrick Hanks, Roser Saurí, Andrew See, Rob Gaizauskas,
  Andrea Setzer, Dragomir Radev, Beth Sundheim, David Day, Lisa Ferro, and
  Marcia Lazo. 2003.
\newblock \href
  {http://ucrel.lancs.ac.uk/publications/cl2003/papers/pustejovsky.pdf} {The
  {TimeBank} corpus}.
\newblock \emph{Proceedings of Corpus Linguistics}.

\bibitem[{Pyysalo et~al.(2007)Pyysalo, Ginter, Heimonen, Björne, Boberg,
  Järvinen, and Salakoski}]{pyysalo2007bioinfer}
Sampo Pyysalo, Filip Ginter, Juho Heimonen, Jari Björne, Jorma Boberg, Jouni
  Järvinen, and Tapio Salakoski. 2007.
\newblock \href {https://doi.org/10.1186/1471-2105-8-50} {{BioInfer}: {A}
  corpus for information extraction in the biomedical domain}.
\newblock \emph{BMC bioinformatics}, 8:50.

\bibitem[{Rajpurkar et~al.(2016)Rajpurkar, Zhang, Lopyrev, and
  Liang}]{rajpurkar-etal-2016-squad}
Pranav Rajpurkar, Jian Zhang, Konstantin Lopyrev, and Percy Liang. 2016.
\newblock \href {https://doi.org/10.18653/v1/D16-1264} {{SQ}u{AD}: 100,000+
  questions for machine comprehension of text}.
\newblock In \emph{Proceedings of EMNLP}, pages 2383--2392.

\bibitem[{Ritter et~al.(2012)Ritter, Mausam, Etzioni, and
  Clark}]{ritter2012open}
Alan Ritter, Mausam Mausam, Oren Etzioni, and Sam Clark. 2012.
\newblock \href {https://doi.org/10.1145/2339530.2339704} {Open domain event
  extraction from twitter}.
\newblock \emph{Proceedings of the ACM SIGKDD}, 1104-1112.

\bibitem[{Sims et~al.(2019)Sims, Park, and Bamman}]{sims-etal-2019-literary}
Matthew Sims, Jong~Ho Park, and David Bamman. 2019.
\newblock \href {https://doi.org/10.18653/v1/P19-1353} {Literary event
  detection}.
\newblock In \emph{Proceedings of ACL}, pages 3623--3634.

\bibitem[{Song et~al.(2015)Song, Bies, Strassel, Riese, Mott, Ellis, Wright,
  Kulick, Ryant, and Ma}]{song-etal-2015-light}
Zhiyi Song, Ann Bies, Stephanie Strassel, Tom Riese, Justin Mott, Joe Ellis,
  Jonathan Wright, Seth Kulick, Neville Ryant, and Xiaoyi Ma. 2015.
\newblock \href {https://doi.org/10.3115/v1/W15-0812} {From light to rich
  {ERE}: Annotation of entities, relations, and events}.
\newblock In \emph{Proceedings of the The 3rd Workshop on {EVENTS}: Definition,
  Detection, Coreference, and Representation}, pages 89--98.

\bibitem[{Swayamdipta et~al.(2017)Swayamdipta, Thomson, Dyer, and
  Smith}]{swayamdipta2017frame}
Swabha Swayamdipta, Sam Thomson, Chris Dyer, and Noah~A Smith. 2017.
\newblock \href {https://arxiv.org/abs/1706.09528} {Frame-semantic parsing with
  softmax-margin segmental rnns and a syntactic scaffold}.
\newblock \emph{arXiv preprint arXiv:1706.09528}.

\bibitem[{Thompson et~al.(2009)Thompson, Iqbal, McNaught, and
  Ananiadou}]{thompson2009construction}
Paul Thompson, Syed Iqbal, John McNaught, and Sophia Ananiadou. 2009.
\newblock \href {https://doi.org/10.1186/1471-2105-10-349} {Construction of an
  annotated corpus to support biomedical information extraction}.
\newblock \emph{BMC bioinformatics}, 10:349.

\bibitem[{Walker et~al.(2006)Walker, Strassel, Medero, and
  Maeda}]{walker2006ace}
Christopher Walker, Stephanie Strassel, Julie Medero, and Kazuaki Maeda. 2006.
\newblock \href {https://catalog.ldc.upenn.edu/LDC2006T06} {{ACE} 2005
  multilingual training corpus}.
\newblock \emph{Linguistic Data Consortium, Philadelphia}, 57.

\bibitem[{Wang et~al.(2019{\natexlab{a}})Wang, Hula, Xia, Pappagari, McCoy,
  Patel, Kim, Tenney, Huang, Yu, Jin, Chen, Van~Durme, Grave, Pavlick, and
  Bowman}]{wang-etal-2019-tell}
Alex Wang, Jan Hula, Patrick Xia, Raghavendra Pappagari, R.~Thomas McCoy, Roma
  Patel, Najoung Kim, Ian Tenney, Yinghui Huang, Katherin Yu, Shuning Jin,
  Berlin Chen, Benjamin Van~Durme, Edouard Grave, Ellie Pavlick, and Samuel~R.
  Bowman. 2019{\natexlab{a}}.
\newblock \href {https://doi.org/10.18653/v1/P19-1439} {{C}an {Y}ou {T}ell {M}e
  {H}ow to {G}et {P}ast {S}esame {S}treet? {S}entence-{L}evel {P}retraining
  {B}eyond {L}anguage {M}odeling}.
\newblock In \emph{Proceedings of ACL}, pages 4465--4476.

\bibitem[{Wang et~al.(2019{\natexlab{b}})Wang, Han, Liu, Sun, and
  Li}]{wang-etal-2019-adversarial-training}
Xiaozhi Wang, Xu~Han, Zhiyuan Liu, Maosong Sun, and Peng Li.
  2019{\natexlab{b}}.
\newblock \href {https://doi.org/10.18653/v1/N19-1105} {Adversarial training
  for weakly supervised event detection}.
\newblock In \emph{Proceedings of NAACL}, pages 998--1008.

\bibitem[{Wang et~al.(2019{\natexlab{c}})Wang, Wang, Han, Liu, Li, Li, Sun,
  Zhou, and Ren}]{wang-etal-2019-hmeae}
Xiaozhi Wang, Ziqi Wang, Xu~Han, Zhiyuan Liu, Juanzi Li, Peng Li, Maosong Sun,
  Jie Zhou, and Xiang Ren. 2019{\natexlab{c}}.
\newblock \href {https://doi.org/10.18653/v1/D19-1584} {{HMEAE}: {H}ierarchical
  modular event argument extraction}.
\newblock In \emph{Proceedings of EMNLP-IJCNLP}, pages 5777--5783.

\bibitem[{Wolf et~al.(2019)Wolf, Debut, Sanh, Chaumond, Delangue, Moi, Cistac,
  Rault, Louf, Funtowicz, and Brew}]{Wolf2019HuggingFacesTS}
Thomas Wolf, Lysandre Debut, Victor Sanh, Julien Chaumond, Clement Delangue,
  Anthony Moi, Pierric Cistac, Tim Rault, R'emi Louf, Morgan Funtowicz, and
  Jamie Brew. 2019.
\newblock \href {https://arxiv.org/abs/1910.03771} {{H}ugging{F}ace's
  {T}ransformers: {S}tate-of-the-art natural language processing}.
\newblock \emph{arXiv preprint arXiv:1910.03771}.

\bibitem[{Yan et~al.(2019)Yan, Jin, Meng, Guo, and Cheng}]{yan-etal-2019-event}
Haoran Yan, Xiaolong Jin, Xiangbin Meng, Jiafeng Guo, and Xueqi Cheng. 2019.
\newblock \href {https://doi.org/10.18653/v1/D19-1582} {Event detection with
  multi-order graph convolution and aggregated attention}.
\newblock In \emph{Proceedings of EMNLP-IJCNLP}, pages 5766--5770.

\bibitem[{Yang et~al.(2003)Yang, Chua, Wang, and Koh}]{yang2003structured}
Hui Yang, Tat-Seng Chua, Shuguang Wang, and Chun-Keat Koh. 2003.
\newblock \href {https://doi.org/10.1145/860435.860444} {Structured use of
  external knowledge for event-based open domain question answering}.
\newblock In \emph{Proceedings of SIGIR}, pages 33--40.

\bibitem[{Yang et~al.(2019)Yang, Zhou, He, and Zhang}]{yang2019interpretable}
Yang Yang, Deyu Zhou, Yulan He, and Meng Zhang. 2019.
\newblock \href {https://doi.org/10.18653/v1/D19-1017} {Interpretable relevant
  emotion ranking with event-driven attention}.
\newblock In \emph{Proceedings of EMNLP-IJCNLP}, pages 177--187.

\bibitem[{Yang et~al.(2018)Yang, Qi, Zhang, Bengio, Cohen, Salakhutdinov, and
  Manning}]{yang-etal-2018-hotpotqa}
Zhilin Yang, Peng Qi, Saizheng Zhang, Yoshua Bengio, William Cohen, Ruslan
  Salakhutdinov, and Christopher~D. Manning. 2018.
\newblock \href {https://doi.org/10.18653/v1/D18-1259} {{H}otpot{QA}: A dataset
  for diverse, explainable multi-hop question answering}.
\newblock In \emph{Proceedings of EMNLP}, pages 2369--2380.

\bibitem[{Yao et~al.(2019)Yao, Ye, Li, Han, Lin, Liu, Liu, Huang, Zhou, and
  Sun}]{yao-etal-2019-docred}
Yuan Yao, Deming Ye, Peng Li, Xu~Han, Yankai Lin, Zhenghao Liu, Zhiyuan Liu,
  Lixin Huang, Jie Zhou, and Maosong Sun. 2019.
\newblock \href {https://doi.org/10.18653/v1/P19-1074} {{D}oc{RED}: A
  large-scale document-level relation extraction dataset}.
\newblock In \emph{Proceedings of ACLs}, pages 764--777.

\bibitem[{Zeiler(2012)}]{Zeiler2012ADADELTAAA}
Matthew~D. Zeiler. 2012.
\newblock \href {https://arxiv.org/abs/1212.5701} {{ADADELTA}: {A}n adaptive
  learning rate method}.
\newblock \emph{ArXiv}, abs/1212.5701.

\bibitem[{Zeng et~al.(2018)Zeng, Feng, Ma, Wang, Yan, Shi, and
  Zhao}]{zeng2018scale}
Ying Zeng, Yansong Feng, Rong Ma, Zheng Wang, Rui Yan, Chongde Shi, and Dongyan
  Zhao. 2018.
\newblock \href
  {https://www.aaai.org/ocs/index.php/AAAI/AAAI18/paper/viewFile/16119/16173}
  {Scale up event extraction learning via automatic training data generation}.
\newblock In \emph{Proceedings of AAAI}.

\bibitem[{Zhao et~al.(2018)Zhao, Jin, Wang, and
  Cheng}]{zhao-etal-2018-document}
Yue Zhao, Xiaolong Jin, Yuanzhuo Wang, and Xueqi Cheng. 2018.
\newblock \href {https://doi.org/10.18653/v1/P18-2066} {Document embedding
  enhanced event detection with hierarchical and supervised attention}.
\newblock In \emph{Proceedings of ACL}, pages 414--419.

\end{thebibliography}
\bibliographystyle{acl_natbib}

\appendix
\section{Hyperparameter Settings and Training Details}
\label{app:hyper}
In this section, we introduce the hyperparameter settings and training details of various ED models that we implemented for experiments.

\subsection{BERT-based Models}
For both \textbf{DMBERT} and \textbf{BERT-CRF}, we use the BERT$_{\small \texttt{BASE}}$ model and the released pre-trained checkpoints\footnote{\url{https://github.com/google-research/bert}}, and implement them with HuggingFace's Transformers library~\cite{Wolf2019HuggingFacesTS}. The two models are both trained with the AdamW\footnote{\url{https://www.fast.ai/2018/07/02/adam-weight-decay/\#adamw}} optimizer and share most of the hyperparameters. Their hyperparameters are shown in Table~\ref{table:hpbert}.

For the \textbf{DMBERT} model, we insert special tokens (\texttt{[unused0]} and \texttt{[unused1]}) around the trigger candidates to indicate their positions and use a much larger batch size, hence the results are higher than the original implementation~\cite{wang-etal-2019-adversarial-training}. 

For the \textbf{BERT+CRF} model, we use the widely-used ``BIO'' tagging schema, where ``B-EventType'', ``I-EventType'' and ``O'' stand for ``Begin Event Type'', ``Inside Event Type'' and ``Others'' respectively.

\begin{table}[htbp]
\small
\centering
\scalebox{0.9}{
\begin{tabular}{l|c}
\toprule
Learning Rate    & $5\times10^{-5}$ \\
Adam $\epsilon$ & $1\times10^{-8}$ \\
Warmup Rate & $0.0$     \\
DMBERT Batch Size      & $336$    \\
BERT-CRF Batch Size   & $256$ \\
DMBERT Validation Steps & $500$ \\
BERT-CRF Validation Steps on MAVEN & $100$ \\
BERT-CRF Validation Steps on ACE 2005& $50$ \\
\bottomrule
\end{tabular}}
\caption{Hyperparameter settings for the BERT-based models.}
\label{table:hpbert}
\end{table}

\subsection{MOGANED Model}
\textbf{MOGANED} model is implemented by ourselves since the official codes are not released. Compared with the original paper, our reproduction uses Adam optimizer and does not use the L2 norm, while other model details are the same as~\newcite{yan-etal-2019-event}. We set most hyperparameters same as~\newcite{yan-etal-2019-event} but the hyperparameter $\lambda$ to be $1$ rather than $5$ since we find it can achieve better performances on both datasets. For the hyperparameters not mentioned in the original paper, we tune them manually. All hyperparameters are shown in Table~\ref{table:hpmoganed}.

\begin{table}[htbp]
\centering
\small
\scalebox{0.9}{
\begin{tabular}{l|c}
\toprule
$K$ & $3$     \\
$\lambda$ & $1$     \\
Batch Size      & $30$    \\
Leaky Alpha & $0.2$     \\
Dropout Rate & $0.3$     \\
Learning Rate    & $1\times10^{-3}$ \\
Dimension of Pos-Tag Feature & $50$     \\
Dimension of NER-Tag Feature & $50$     \\
Dimension of Word Embedding  & $100$     \\
Dimension of Position Embedding  & $50$     \\
Dimension of Hidden Feature  & $100$     \\
Dimension of Graph Feature  & $150$     \\
Dimension of $W_{att}$ Feature  & $100$     \\
Dimension of Aggregation Feature  & $100$     \\
\bottomrule
\end{tabular}}
\caption{Hyperparameter settings for MOGANED.}
\label{table:hpmoganed}
\end{table}

\subsection{DMCNN model}
\textbf{DMCNN} model is implemented by ourselves since the official codes are not released. Compared with ~\newcite{chen2015event}, we use Adam optimizer instead of the ADADELTA~\cite{Zeiler2012ADADELTAAA} optimizer. We set all the hyperparameters the same as \newcite{chen2015event} except the word embedding dimension and learning rate, which are not mentioned in the original paper. As the pre-trained word embeddings used in the original paper are not publicly released, we use the pre-trained word embeddings released by \newcite{chen-etal-2018-collective} instead. The hyperparameters are shown in Table~\ref{table:hpdmcnn}.

\begin{table}[htbp]
\small
\centering
\scalebox{0.9}{
\begin{tabular}{l|c}
\toprule
Batch Size      & $170$    \\
Dropout Rate   & $0.5$    \\
Learning Rate   & $1\times10^{-3}$    \\
Adam $\epsilon$     & $1\times10^{-8}$   \\
Kernel Size     & $3$      \\
Dimension of PF    & $5$   \\
Number of Feature Map   & $200$    \\
Dimension of Word Embedding  & $100$    \\
\bottomrule
\end{tabular}}
\caption{Hyperparameter settings for DMCNN.}
\label{table:hpdmcnn}
\vspace{-2em}
\end{table}

\subsection{BiLSTM-based Models}
For both \textbf{BiLSTM} and \textbf{BiLSTM-CRF}, we use the pre-trained word embeddings released by~\newcite{chen-etal-2018-collective} and train them with the Adam~\cite{adam} optimizer. Similar with \textbf{BERT-CRF}, we use ``BIO'' tagging schema in \textbf{BiLSTM-CRF}. Their hyperparameters are shown in Table~\ref{table:hpbilstm}.

\begin{table}[htbp]
\small
\centering
\scalebox{0.9}{
\begin{tabular}{@{}l|c@{}}
\toprule
Batch Size      & $200$    \\
Dropout Rate   & $0.3$    \\
Learning Rate   & $1\times10^{-3}$     \\
Adam $\epsilon$     & $1\times10^{-8}$   \\
Dimension of Hidden Layers    & $200$   \\
Dimension of Word Embedding  & $100$    \\
\bottomrule
\end{tabular}}
\caption{Hyperparameter settings for the BiLSTM-based models.}
\label{table:hpbilstm}
\vspace{-2em}
\end{table}

\subsection{Overall Training Details}
For reproducibility, we report the training details of various models in this section. Table~\ref{table:traindetails} shows the used computing infrastructures, the numbers of model parameters as well as the average running time of various models.

We mostly follow the original hyperparameter settings but also manually tune some hyperparameters. We select the models with the F-1 scores on the development sets of the both datasets. The validation performances of various models are shown in Table~\ref{table:val}.

\begin{table}[th]
\centering
\small
\scalebox{0.8}{
\begin{tabular}{l|c|c|cc}
\toprule
\multirow{2}{*}{\textbf{Method}} & \multirow{2}{*}{\begin{tabular}[c]{@{}c@{}}\textbf{Computing} \\ \textbf{Infrastructure}\end{tabular}} & \multicolumn{1}{l|}{\multirow{2}{*}{\textbf{\#para.}}} & \multicolumn{2}{c}{\textbf{Runtime}} \\ \cmidrule{4-5} 
                        &                                                                                     & \multicolumn{1}{l|}{}                             & \textbf{ACE 2005}  & \textbf{MAVEN}   \\ \midrule
DMCNN                   & $1\times$ RTX 2080 Ti  & $2$M &    $3$~min  & $5.5$~min \\
BiLSTM                  & $1\times$ RTX 2080 Ti  & $2$M & $18$~min & $29$~min  \\
BiLSTM+CRF              & $1\times$ RTX 2080 Ti  & $3$M &       $21$~min & $67$~min  \\
MOGANED                 & $1\times$ RTX 2080 Ti   & $40$M &     $55$~min & $90$~min  \\
DMBERT                  & $8\times$ RTX 2080 Ti & $110$M &  $110$~min & $201$~min \\
BERT+CRF                & $1\times$ RTX 2080 Ti & $110$M & $32$~min &  $97$~min  \\ \bottomrule
\end{tabular}
}
\caption{Training details of various models, including the computing infrastructures, the numbers of parameters, and the average runtimes.}
\label{table:traindetails}
\end{table}

\begin{table}[h]
\centering
\small
\scalebox{0.9}{
\begin{tabular}{l|ccc|ccc}
\toprule
\multirow{2}{*}{\textbf{Method}} & \multicolumn{3}{c|}{\textbf{ACE 2005}} & \multicolumn{3}{c}{\textbf{MAVEN}} \\ \cmidrule{2-7} 
                        & \textbf{P} & \textbf{R} & \textbf{F-1} & \textbf{P}  & \textbf{R}   & \textbf{F-1}    \\ \midrule
DMCNN                   & $73.3$    & $53.5$  & $61.8$  & $66.5$ & $55.5$    & $60.5$    \\
BiLSTM                  & $72.3$    & $67.6$  & $69.8$  & $60.3$ & $66.9$   & $63.4$    \\
BiLSTM+CRF              & $75.9$  & $60.8$  & $67.5$    & $64.1$ & $64.6$     & $64.3$  \\
MOGANED                 & $72.4$  & $66.2$  & $69.1$    & $63.7$     & $63.7$     & $63.7$    \\
DMBERT                  & $71.4$  & $72.4$  & $71.9$     & $64.6$  & $70.1$  & $67.2$    \\
BERT+CRF                & $75.4$  & $76.8$   & $76.1$   & $65.7$  & $68.8$  & $67.2$    \\ \bottomrule
\end{tabular}
}
\caption{Validation performance of various models.}
\label{table:val}
\vspace{-1.5em}
\end{table}

\section{Event Type Mapping for ACE and MAVEN}
\label{app:acemapping}
In Table~\ref{tab:TypeMapping}, we present the event type mapping between parts of ACE 2005 and MAVEN event types, which is used in the data augmentation experiments in Section~\ref{sec:transfer}.

\begin{table}[!htp]
\small
\scalebox{0.8}{
\begin{tabular}{l|c}
\toprule
\textbf{ACE Types}                   & \textbf{MAVEN Types}                                                                                                                                                                                                                                  \\ \midrule
\texttt{Injure}             & \texttt{Bodily\_harm}                                                                                                                                                                                                                        \\ \hline
\texttt{Die}                & \texttt{Death}                                                                                                                                                                                                                               \\ \hline
\texttt{Transport}          & \texttt{Traveling}                                                                                                                                                                                                                           \\ \hline
\texttt{Transfer-Ownership} & \begin{tabular}[c]{@{}c@{}}\texttt{Getting},\texttt{Receving},\\ \texttt{Commerce\_buy}, \texttt{Giving}, \\ \texttt{Submitting\_documents}, \texttt{Supply},\\ \texttt{Commerce\_sell}, \texttt{Renting}, \\ \texttt{Exchange}\end{tabular} \\ \hline
\texttt{Transfer-Money}     & \begin{tabular}[c]{@{}c@{}}\texttt{Commerce\_pay}, \texttt{Expensiveness},\\ \texttt{Earnings\_and\_losses}\end{tabular}                                                                                                                     \\ \hline
\texttt{Attack}             & \texttt{Attack}                                                                                                                                                                                                                              \\ \hline
\texttt{Demonstrate}        & \texttt{Protest}                                                                                                                                                                                                                             \\ \hline
\texttt{Meet}               & \texttt{Come\_together}, \texttt{Social\_event}                                                                                                                                                                                              \\ \hline
\texttt{Phone-Write}        & \texttt{Communication}, \texttt{Telling}                                                                                                                                                                                                     \\ \hline
\texttt{Arrest-Jail}        & \texttt{Arrest}, \texttt{Prison}                                                                                                                                                                                                             \\ \hline
\texttt{Extradite}          & \texttt{Extradition}                                                                                                                                                                                                                         \\ \hline
\texttt{Trial-Hearing}      & \texttt{Justifying}                                                                                                                                                                                                                          \\ \bottomrule
\end{tabular}
}
\caption{Mapping between parts of ACE 2005 event types and MAVEN event types.}
\label{tab:TypeMapping}
\vspace{-2em}
\end{table}

\section{Hierarchical Event Type Schema}
\label{app:hierschema}
\begin{figure*}[htp]
\centering
\includegraphics[width = 0.9\textwidth]{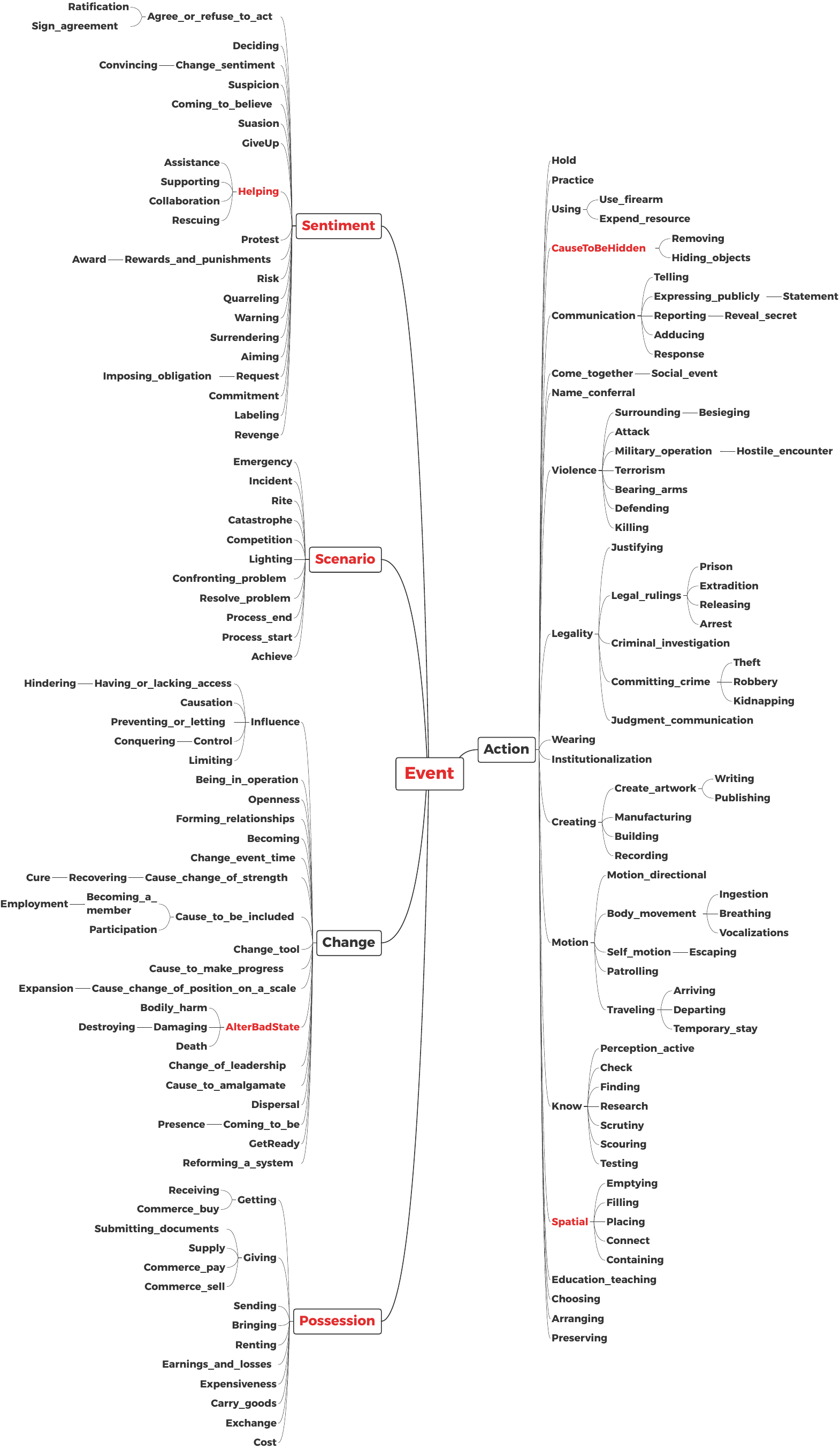}
\caption{The hierarchical event type schema used in MAVEN. The \textcolor{red}{red} labels are virtual event types without annotated instances.}
\label{fig:schema}
\end{figure*}

We present the tree-structure hierarchical event type schema used by MAVEN in Figure~\ref{fig:schema}. The eight red types are virtual types without annotated instances, which are only used for organizing similar event types together. The virtual types do not participate in classification for all the models and when we say we have $168$ event types we do not take them into account.

\section{Event Types and their corresponding frames}
\label{app:frames}
As stated in Section 3.1, we manually induce $168$ event types from the $598$ FrameNet event-related frames. We present the mapping between the event types and frames in Table~\ref{table:frames} to help understand our event schema construction process. Note that the shown mapping is not a strict mapping, i.e., the semantic coverage of a MAVEN event type may be larger than the union of its corresponding frames.

\clearpage
\onecolumn
\begin{center}
\begin{small}
\begin{longtable}{l|c}
\toprule 
\textbf{Event Type}                             & \textbf{Corresponding Frame(s)}                                                                                                                                                                                                                                   \\\midrule
\texttt{Know}                                      & \texttt{Becoming\_aware}                                                                                                                                                                                                                                                               \\ \midrule
\texttt{Warning}                                   & \texttt{Warning}                                                                                                                                                                                                                                                                       \\ \midrule
\texttt{Catastrophe}                               & \texttt{Catastrophe}                                                                                                                                                                                                                                                                   \\ \midrule
\texttt{Placing}                                   & \texttt{Placing\_scenario}, \texttt{Placing}, \texttt{Being\_located}                                                                                                                                                                                                                  \\
\end{longtable}
\end{small}
\end{center}
\clearpage
\twocolumn

\end{document}